\documentclass[10pt,twocolumn,letterpaper]{article}

\usepackage{cvpr}              

\definecolor{darkgreen}{rgb}{0, 0.5, 0}   
\definecolor{darkred}{rgb}{0.6, 0, 0}     

\usepackage{amsmath,amssymb,amsfonts}
\usepackage{algorithmic}
\usepackage{graphicx}
\usepackage{textcomp}
\usepackage{xcolor}
\usepackage{multirow}  
\usepackage{makecell}  
\usepackage{colortbl}

\usepackage[accsupp]{axessibility}  

\definecolor{cvprblue}{rgb}{0.21,0.49,0.74}
\usepackage[pagebackref,breaklinks,colorlinks,allcolors=cvprblue]{hyperref}

\def\paperID{6895} 
\def\confName{CVPR}
\def\confYear{2026}

\title{Soft Modality-Guided Expert Specialization in MoE-VLMs}

\newcommand{\method}{\textsc{SMoES}}

\author{
    Zi-Hao Bo \quad Yaqian Li* \quad Anzhou Hou \quad Rinyoichi Takezoe \quad Ertao Zhao \\
    Tianxiang Pan \quad Jiale Yan \quad Mo Guang \quad Kaiwen Long \\
    Li Auto Inc.\\
    {\tt\small * liyaqian@lixiang.com}
}

\begin{document}

\maketitle

\begin{abstract}
    Mixture-of-Experts (MoE) has become a prevalent backbone for large vision-language models (VLMs), yet how modality-specific signals should guide expert routing remains under-explored.
    Existing routing strategies are either hand-crafted or modality-agnostic, relying on idealized priors that ignore the layer-dependent modality fusion patterns in MoE-VLMs and provide little guidance for expert specialization.
    We propose Soft Modality-guided Expert Specialization (SMoES), which consists of dynamic soft modality scores that capture layer-dependent fusion patterns, an expert binning mechanism aligned with expert-parallel deployment, and an inter-bin mutual information regularization that encourages coherent modality specialization.
    Our method leverages attention-based or Gaussian-statistics modality scores to optimize mutual information regularization.
    Experiments across four MoE-based VLMs and 16 benchmarks demonstrate improvement on both effectiveness and efficiency: 0.9\% and 4.2\% average gain on multimodal and language-only tasks, 56.1\% reduction in EP communication overhead, and 12.3\% throughput improvement under realistic deployment.
    These results validate that aligning routing with modality-aware expert specialization unlocks MoE-VLM capacity and efficiency.
\end{abstract}    
\section{Introduction}
\label{sec:intro}

The Mixture-of-Experts (MoE) architecture~\cite{shazeer2017moe,lepikhin2020gshard,fedus2022switch-transformers,guo2025deepseek-r1} has emerged as a cornerstone for modern large vision-language models (VLMs). 
Its principle of conditional computation allows for a massive expansion of model capacity with only a modest increase in the per-token computational budget by routing inputs to specialized expert sub-networks~\cite{shazeer2017moe,kudugunta2021task-moe}. 
This paradigm is particularly well-suited for fusing heterogeneous modalities. 
Consequently, leading systems like DeepSeek-VL2~\cite{wu2024deepseek-vl2}, Kimi-VL~\cite{team2025kimi-vl}, GLM-4.5V~\cite{hong2025glm4.5v}, and InternVL-3.5~\cite{wang2025internvl3.5} have adopted MoE to achieve better performance while maintaining tractable inference costs. 
Despite this widespread adoption, a fundamental question remains under-explored: how do modality-specific signals (vision vs. text) interact with the expert routing mechanism, and can it be optimized to enhance both effectiveness and efficiency?

\begin{figure}[t]
  \centering
  \includegraphics[width=0.99\columnwidth]{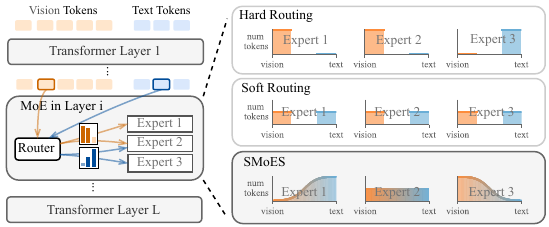}
  \vspace{-0.3em}
  \caption{Comparison of routing strategies in MoE-VLMs. 
      Hard routing enforces strict modality-expert separation, while soft routing allows arbitrary, uncontrolled mixing. 
      SMoES uses soft modality scores to encourage dynamic expert specialization.
  }
  \label{fig:overview}
  \vspace{-1.2em}
\end{figure}

Mainstream MoE-VLMs route tokens using either hard or soft routing paradigms (\cref{fig:overview}).
Hard routing~\cite{bao2022vlmo,wang2023beit3} pre-assigns experts to a specific modality, creating sharp specialization at the cost of rigid boundaries that hinder adaptation to cross-modal features and ignore the natural blending of representations across layers.
In contrast, soft routing~\cite{mustafa2022limoe,wu2024deepseek-vl2,team2025kimi-vl,hong2025glm4.5v}, the prevailing paradigm, allows experts to process any token.
Yet, these methods often rely on heuristic priors or auxiliary losses disconnected from the evolving modality distributions, resulting in either over-mixing or under-specialization.
Hybrid approaches~\cite{bao2022vlmo,wang2023beit3,wang2025moiie} partition experts into modality-specific (hard-routed) and shared (soft-routed) groups, but this partitioning is typically hand-crafted and layer-agnostic, failing to align with feature evolution.
Fundamentally, these paradigms are guided by idealized assumptions or uncontrolled mixing rather than the dynamic, data-driven geometry of modality distributions, offering little guidance on how expert specialization should evolve with network depth.

\begin{figure}[t]
  \centering
  \includegraphics[width=0.99\columnwidth]{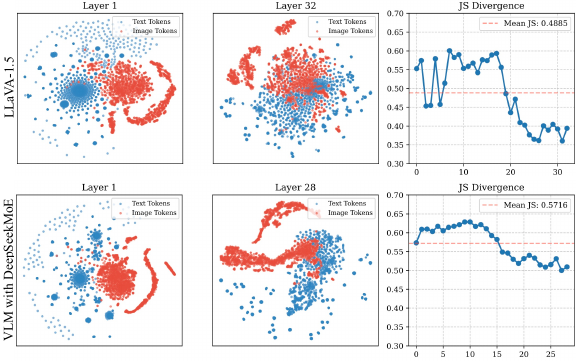}
  \vspace{-0.5em}
  \caption{Modality fusion pattern.
      Token feature distributions vary across models and layers.
      Text tokens are divided into several semantic clusters, while vision tokens are roughly separated into content and padding clusters.
      The last column shows the JS divergence between vision and text tokens across all layers.
  }
  \label{fig:modality-fusion}
  \vspace{-1.7em}
\end{figure}

We analyze modality fusion patterns of token features across layers in LLaVA-1.5~\cite{liu2024llava-1.5} and a DeepSeekMoE-based~\cite{dai2024deepseekmoe} VLM, as shown in \cref{fig:modality-fusion}.
The analysis shows that modality fusion in MoE-VLMs is highly heterogeneous, with no clear boundary between modality-specific and cross-modal states.
At a macro level, fusion patterns vary across models and layers, as reflected by distinct Jensen–Shannon (JS) divergence trajectories.
At a micro level, even within the same layer and modality, some tokens remain modality-specific while others become cross-modal.
This multi-scale heterogeneity indicates that rigid routing assumptions---either enforcing hard separation or imposing uniform mixing---are misaligned with how modalities actually interact across depth.
Therefore, guided by signals that respect the evolving modality structure, the expert pool in MoE has the potential to naturally enable dynamic modality specialization.

Beyond effectiveness, modality specialization is also closely tied to the inference efficiency of MoE-VLMs.
Vision and text tokens differ sharply in both quantity and information density.
Vision tokens often dominate the sequence length but typically carry lower information density due to spatial redundancy, whereas text tokens are fewer yet more semantically concentrated~\cite{chen2024fast-v,jin2024chat-univi,meng2024deepstack}.
In real-world scenarios, inputs may even be text-only.
This asymmetry poses two efficiency challenges.
First, standard routing objectives coupled with vanilla load balancing tend to allocate most experts to the dominant yet low-information-density vision modality, hindering optimal specialization.
Second, under expert parallelism (EP)~\cite{liu2024deepseek-v3}---a common deployment strategy---modality-agnostic routing scatters tokens across devices, inflating inter-communication overhead.
Carefully-designed modality specialization addresses both issues: by establishing clear expert–modality affinities and scheduling tokens accordingly, aligned experts can be co-located on the same device, reducing communication while preserving balanced computation and capacity.

Motivated by these observations, we propose Soft Modality-guided Expert Specialization (\method{}), which addresses these challenges through three key components.
First, we introduce dynamic soft modality scores through two complementary estimators---attention-accumulated and Gaussian-statistics---that enable adaptive expert selection aligned with actual fusion states rather than hard boundaries.
Second, we introduce expert binning---partitioning experts into coherent groups---which serves as the structural foundation for modality specialization and enables natural device placement units for modality-aware expert parallelism deployment.
Third, to drive effective modality specialization, we introduce a mutual information (MI) objective that encourages different expert bins to specialize on distinct modality patterns.

In summary, we make three key contributions:
(1) Dynamic soft modality scores---Gaussian-statistics and attention-accumulated estimators---that align routing with the evolving modality fusion patterns across layers.
(2) An expert binning mechanism coupled with an MI-driven specialization objective that promotes coherent modality grouping for both effectiveness and deployment efficiency.
(3) Extensive experiments across four MoE-VLM backbones and 16 benchmarks demonstrating SOTA performance in accuracy and efficiency, including reduced EP communication overhead in realistic deployment.

\section{Related Works}
\label{sec:related-work}

\begin{figure*}[t]
    \centering
    \includegraphics[width=0.99\textwidth]{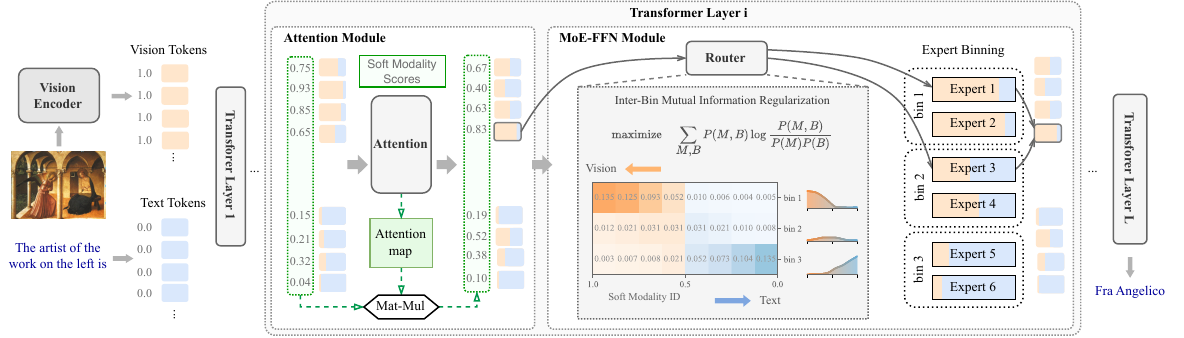}
    \vspace{-0.5em}
    \caption{Overview of \method{}. 
        \textbf{Soft Modality Scores}: Tokens start with hard modality IDs (0/1 for vision/text) and are progressively refined into layer-wise soft scores $M\in[0,1]$. 
        We estimate them via (i) attention-accumulated (depicted here) or (ii) Gaussian-statistics estimation (detailed in \cref{fig:gaussian-score}). 
        \textbf{Expert Binning}: experts learn bin-level modality preference for deployment efficiency. 
        \textbf{Mutual Information} Regularization: maximizes mutual information between soft modality scores ($M$) and bins ($B$), inducing modality preference.
    }
    \label{fig:method}
    \vspace{-1.3em}
\end{figure*}

\subsection{Mixture-of-Experts}

Sparsely-gated MoE architectures enable conditional computation through sparse activation and efficient gating~\cite{shazeer2017moe,fedus2022switch-transformers,lepikhin2020gshard}. 
Recent works like DeepSeek-V3 further improve routing efficiency, load balancing, and communication performance~\cite{puigcerver2024soft-moe,zhou2022ec-moe,liu2024deepseek-v3}. 
Mutual-information (MI) constraints have been explored to regulate expert selection.
For expert-task alignment, Mod-Squad and MTHL maximize MI between tasks and experts~\cite{chen2023mod-squad,chen2023mthl}.
For expert-token alignment, ModuleFormer and CoMoE maximize MI between tokens and modules through entropy balancing or contrastive learning~\cite{shen2023module-former,feng2025co-moe}.
However, these methods overlook modality differentiation in multimodal models.
Recent VLMs increasingly adopt MoE backbones~\cite{wu2024deepseek-vl2,team2025kimi-vl,hong2025glm4.5v,guo2025seed1.5-vl,wang2025internvl3.5,wang2025step-3}.
Related approaches also include LoRA-based expert expansion~\cite{wu2024mole,chen2024llava-mole,shen2024mome} and converting dense VLMs to sparse MoE architectures~\cite{lin2024moe-llave,zheng2024lte,jing2025evo-moe}.

\subsection{Modality Specialization in MoE-based VLMs}

For modality-aware routing, existing methods can be categorized by their degree of expert-modality coupling.
\textbf{Hard routing} assigns experts exclusively to single modalities~\cite{shen2023vl-moe,liang2025mot,li2025uni-moe,luo2025mono-internvl,deng2025bagel,ai2025ming-omni,lin2024moma}, achieving strong specialization but sacrificing flexibility across modalities.
\textbf{Hybrid routing} allows certain experts to handle multiple modalities while others remain modality-specific~\cite{bao2022vlmo,wang2023beit3,wang2025moiie}, yet relies on manual partitioning misaligned with actual fusion dynamics.
\textbf{Naive soft routing}~\cite{mustafa2022limoe,chen2024eve} permits flexible expert mixing without explicit modality guidance, leaving specialization to implicit optimization.
\textbf{Smarter soft routing} introduces auxiliary objectives to shape routing: LTDR balances expert load on long-tailed vision distributions~\cite{cai2025ltdr}, and STGC mitigates gradient conflicts within experts~\cite{yang2024stgc}.
SMAR achieves modality differentiation by using KL divergence to regularize routing distributions toward modality-specific patterns~\cite{xia2025smar}.
In contrast to KL-based regularization, our approach employs mutual information constraints between modalities and experts to guide specialization.
Moreover, we introduce layer-adaptive soft modality scores that capture dynamic fusion patterns across layers, and expert binning for efficient distributed deployment.

\subsection{Efficient Deployment of MoE Models}

As MoE models scale, deployment efficiency becomes increasingly critical, with load imbalance and All-to-All communication overhead as key challenges in distributed expert-parallel (EP) settings~\cite{zeng2025efficient-moe}.
To mitigate load imbalance, MoGE enforces balanced expert activation through predefined groups~\cite{tang2025moge}, while Grove-MoE explores heterogeneous expert sizes that dynamically adapt to token complexity~\cite{wu2025grove-moe}.
Complementary strategies address the straggler effect: Expand Drop employs capacity-aware token dropping to limit expert overload~\cite{he2025expand-drop}, and AEP decouples layer execution with asynchronous queuing and adaptive re-batching~\cite{wang2025aep}.
However, for MoE-VLMs, how to leverage modality fusion characteristics to guide expert partitioning and communication strategies under EP deployment remains largely underexplored.

\section{Method}
\label{sec:method}

Modality fusion in MoE-VLMs exhibits layer-varying patterns with smooth token-level transitions, while vision/text asymmetry creates load imbalance and inflates EP communication overhead.
To address these challenges, we introduce \method{}, which integrates three components: soft modality scores capturing continuous token-level modality patterns, expert binning creating modality-aligned partitions, and inter-bin mutual information regularizing coherent specialization, as illustrated in \cref{fig:method}.

\subsection{Preliminaries: MoE in VLM}
\label{sec:preliminaries}

A typical MoE-based VLM consists of a vision encoder, projection layer, and an MoE-augmented LLM backbone.
Within each MoE layer $l$, token features $\mathbf{x}_{ij} \in \mathbb{R}^D$ (where $i$ indexes the sample and $j$ indexes the token) are routed by a gating network to a pool of $N_e$ experts.
The router computes gating scores $g_{ij,e} = \text{softmax}(\mathbf{W}_{\text{gate}} \mathbf{x}_{ij})_e$, and standard top-$k$ routing selects the $k$ highest-scoring experts.
To prevent routing collapse, a load-balancing auxiliary loss~\cite{shazeer2017moe,
fedus2022switch-transformers} is typically applied:
\begin{equation}
\mathcal{L}_{\text{bal}} = \sum_l N_e \sum_{e=1}^{N_e} f_e \, P_e
\end{equation}
where $f_e$ is the fraction of tokens routed to expert $e$, and $P_e$ is the average gating score.

\subsection{Soft Modality Scores}

The multi-scale heterogeneity of modality fusion (\cref{fig:modality-fusion}) reveals smooth, layer-dependent transitions in token representations rather than fixed modality identities.
Hard modality indicators---binary labels assigned at the input---fail to capture these continuous fusion dynamics.
To guide routing aligned with evolving modality patterns, we introduce \textit{soft modality scores} $M_{ij,m}^{(l)} \in [0,1]$ for each token and modality $m \in \{\text{text}, \text{vision}\}$ at layer $l$, where $\sum_m M_{ij,m}^{(l)} = 1$.
We develop two complementary estimators: an \textit{attention-accumulated score} that captures local cross-token interactions; and a \textit{Gaussian-statistics score} that captures global statistical regularities across the dataset.

\subsubsection{Attention-Accumulated Score}
\label{sec:attn-score}

The attention mechanism provides a natural pathway for modality signal propagation: when a token attends to others, it absorbs their modality characteristics proportionally to attention weights.
We initialize scores at layer 0 with hard modality indicators:
\begin{equation}
M_{ij,m}^{\text{attn},(0)} = \mathbf{1}\{ m = m(\mathbf{x}_{ij}) \}.
\end{equation}
For subsequent layers, we update in two steps.
First, aggregate attended tokens' scores using attention weights:
\begin{equation}
\tilde{M}_{ij,m}^{\text{attn},(l)} = \sum_{j'=1}^{J} \text{Attn}_{j,j'}^{(l)} \cdot M_{ij',m}^{\text{attn},(l)}
\end{equation}
where $\text{Attn}^{(l)}$ is the layer $l$ attention matrix averaged across heads, and $J$ is the sequence length.
Second, combine aggregated and input scores via residual-weighted update:
\begin{equation}
M_{ij,m}^{\text{attn},(l+1)} = \frac{\|\mathbf{x}_{\text{attn},ij}^{(l)}\| \cdot \tilde{M}_{ij,m}^{\text{attn},(l)} + \|\mathbf{x}_{ij}^{(l)}\| \cdot M_{ij,m}^{\text{attn},(l)}}{\|\mathbf{x}_{\text{attn},ij}^{(l)}\| + \|\mathbf{x}_{ij}^{(l)}\|}
\end{equation}
This mirrors the residual structure of Transformers ($\mathbf{x}^{(l+1)} = \mathbf{x}^{(l)} + \mathbf{x}_{\text{attn}}^{(l)}$), where feature norms weight the contributions of attention and residual paths.

\subsubsection{Gaussian-Statistics Score}
\label{sec:gaussian-score}

Complementary to the attention-accumulated score's local, sequence-specific perspective, we propose the \textit{Gaussian-statistics score}---a global, distribution-based estimator.
The key insight is that tokens from different modalities exhibit distinct feature distributions in the embedding space. 
We exploit these distributional differences to instantaneously infer modality affiliation at each layer, independent of Layer 0 initialization.

\begin{figure}[t]
    \centering
    \includegraphics[width=0.95\columnwidth]{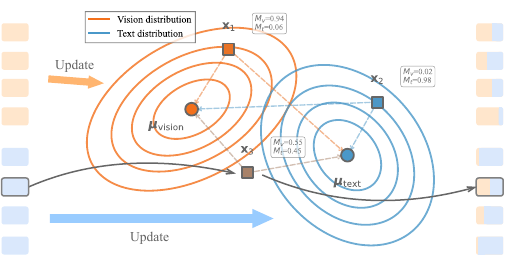}
    \vspace{-0.3em}
    \caption{
        \textbf{Gaussian-statistics score estimation.} Per-modality Gaussian distributions are continuously updated from token batches. For each token, we compute modality affinity as soft scores based on log-likelihood under each distribution.
    }
    \label{fig:gaussian-score}
    \vspace{-1.2em}
\end{figure}

For each layer $l$ and modality $m$, we maintain a Gaussian distribution with diagonal covariance: mean $\boldsymbol{\mu}_m \in \mathbb{R}^D$ and variance $\boldsymbol{\sigma}_m^2 \in \mathbb{R}^D$.
Statistics are updated online via exponential moving average (EMA) variant of Welford's algorithm~\cite{welford1962welford}:
\begin{align}
N_{m,t} &= \beta N_{m,t-1} + |\mathcal{T}_m| \\
\mathbf{S}_{\mu,m,t} &= \beta \mathbf{S}_{\mu,m,t-1} + |\mathcal{T}_m| \cdot \boldsymbol{\mu}_{\mathcal{T}_m} \\
\mathbf{S}_{\sigma^2,m,t} &= \beta \mathbf{S}_{\sigma^2,m,t-1} + \sum_{\mathbf{x} \in \mathcal{T}_m} (\mathbf{x} - \boldsymbol{\mu}_{\mathcal{T}_m})^2 \notag \\
&\quad + (\boldsymbol{\mu}_{\mathcal{T}_m} - \tfrac{\mathbf{S}_{\mu,m,t-1}}{N_{m,t-1}})^2 \cdot \tfrac{\beta N_{m,t-1} |\mathcal{T}_m|}{\beta N_{m,t-1} + |\mathcal{T}_m|}
\end{align}
where $\mathcal{T}_m$ is the set of modality-$m$ tokens in the current batch, $\boldsymbol{\mu}_{\mathcal{T}_m}$ is the batch mean, and $\beta$ is the EMA decay factor. The distribution parameters at time $t$ are: $\boldsymbol{\mu}_m = \mathbf{S}_{\mu,m,t} / N_{m,t}$ and $\boldsymbol{\sigma}_m^2 = \mathbf{S}_{\sigma^2,m,t} / N_{m,t}$.

To infer each token's modality affiliation, we compute its log-likelihood under each distribution:
\begin{equation}
\text{LL}_{ij,m} = -\frac{1}{2} \sum_{d=1}^D \left( \log \sigma_{m,d}^2 + \frac{(x_{ij,d} - \mu_{m,d})^2}{\sigma_{m,d}^2} \right)
\end{equation}
The soft modality score is obtained via temperature-scaled softmax:
\begin{equation}
\label{eq:gaussian-score}
M_{ij,m}^{\text{gauss}} = \frac{\exp(\text{LL}_{ij,m} / \tau)}{\sum_{m' \in \{\text{text}, \text{vision}\}} \exp(\text{LL}_{ij,m'} / \tau)}
\end{equation}
where temperature $\tau$ controls score sharpness.

\subsection{Expert Binning}

In expert-parallel (EP) deployments, modality-agnostic routing scatters tokens across devices, amplifying communication overhead.
To reduce this, we co-locate experts with similar modality preferences on the same device. 
Moreover, real deployments vary in device count, requiring a flexible binning mechanism.

We introduce \textit{expert binning}: at each layer, partition experts into $N_{\text{bins}}$ groups $\mathbf{B} = \{\mathbf{B}_1, \dots, \mathbf{B}_{N_{\text{bins}}}\}$ with $N_B = N_e / N_{\text{bins}}$ experts each, where $N_{\text{bins}}$ can match device count.
By encouraging bins to develop distinct modality preferences through inter-bin MI (detailed below), we enable modality-aligned placement: bins with similar preferences co-locate, reducing communication while balancing load.

To form bins reflecting modality preferences, we adopt \textit{momentum-adaptive binning}.
We track each expert's load from each modality using EMA:
\begin{equation}
\bar{C}_{m,e,t} = \beta \bar{C}_{m,e,t-1} + (1 - \beta) C_{m,e}
\end{equation}
where $C_{m,e} = |\{(i,j) \in \mathcal{T}_m \mid e \in \text{TopK}(g_{ij})\}|$ is the count of modality-$m$ tokens routed to expert $e$.
We compute each expert's text-bias score:
\begin{equation}
f_{\text{spec}}(e) = \frac{\bar{C}_{\text{text},e}}{\bar{C}_{\text{text},e} + \bar{C}_{\text{vision},e}}
\end{equation}
Experts are sorted by $f_{\text{spec}}$ and partitioned into $N_{\text{bins}}$ consecutive bins, grouping experts with similar modality preferences together.

\subsection{Inter-Bin Mutual Information Regularization}

We introduce an \textit{inter-bin MI objective} to drive modality differentiation among expert bins.
The intuition is that high MI between modality $M$ and bin $\mathbf{B}$ means knowing the selected bin provides strong information about token modality, thereby encouraging specialized bins.
Mathematically, we maximize $I(M; \mathbf{B})$.

To compute MI, we first compute the average gating score for each sample $i$, modality $m$, and bin $\mathbf{B}_k$:
\begin{equation}
\bar{S}_{i,m,\mathbf{B}_k} = \frac{\sum_{e \in \mathbf{B}_k} \sum_{j} M_{ij,m} \cdot g_{ij,e}}{N_B \sum_{j} M_{ij,m}}
\end{equation}
where $M_{ij,m}$ is the soft modality score and $g_{ij,e}$ is the gating score. We treat this as an unnormalized distribution and compute the normalized joint probability:
\begin{equation}
P_i(m, \mathbf{B}_k) = \frac{\bar{S}_{i,m,\mathbf{B}_k}}{\sum_{m'} \sum_{k'} \bar{S}_{i,m',\mathbf{B}_{k'}}}
\end{equation}
Marginal probabilities are derived from joint probability, and the MI is:
\begin{equation}
I_i(M; \mathbf{B}) = \sum_m \sum_k P_i(m, \mathbf{B}_k) \log \frac{P_i(m, \mathbf{B}_k)}{P_i(m) \cdot P_i(\mathbf{B}_k)}
\end{equation}
The loss over all layers encourages specialization:
\begin{equation}
\mathcal{L}_{\text{MI}} = -\sum_l \frac{1}{N_{\text{batch}}} \sum_{i=1}^{N_{\text{batch}}} I_i(M; \mathbf{B})
\end{equation}
Operating at the bin level (rather than expert level) naturally aligns with device placement granularity in EP: bins with distinct modality affinities can be co-located on devices, reducing cross-device communication.

\begin{table*}[t]
\centering
\caption{
    Multimodal and language-only results on VLMs based on DeepSeekMoE and OLMoE.
    \textbf{Bold} and \underline{underline}: first and second best performance.
    MSI: Modality Specialization Index. 
    $^\dagger$t/v/s: number of text/vision/shared experts.
    $^*$[a, b]: KL divergence threshold range.
}
\label{tab:main-results-1}
\vspace{-0.3em}
\footnotesize
\setlength{\tabcolsep}{2.pt}
\begin{tabular}{l|c|cccccccccc|c|cccccc|c|c}
\toprule
\textbf{Method} & \textbf{MSI} & \multicolumn{11}{c|}{\textbf{Multimodal Tasks (10)}} & \multicolumn{7}{c|}{\textbf{Language-Only Tasks (6)}} & \textbf{Overall} \\
    & & \rotatebox{70}{$\text{MMMU}^{\text{val}}$} & \rotatebox{70}{$\text{MMMU}^{\text{test}}$} & \rotatebox{70}{GQA} & \rotatebox{70}{POPE} & \rotatebox{70}{SQA-IMG} & \rotatebox{70}{TextVQA} & \rotatebox{70}{MME} & \rotatebox{70}{MMB} & \rotatebox{70}{MMB-CN} & \rotatebox{70}{VQAv2} & \rotatebox{70}{\textbf{Avg}} & \rotatebox{70}{MMLU} & \rotatebox{70}{HellaSwag} & \rotatebox{70}{ARC-C} & \rotatebox{70}{ARC-E} & \rotatebox{70}{GSM8k} & \rotatebox{70}{TruthfulQA} & \rotatebox{70}{\textbf{Avg}} & \rotatebox{70}{\textbf{Avg}} \\
\midrule

\multicolumn{21}{c}{\textit{VLM based on DeepSeek-MoE (A3B/16B, top-6/64 experts)}} \\
\midrule

\textbf{No Specialization} & .177 & 31.9 & 30.5 & 59.5 & 84.9 & 68.0 & 56.5 & \underline{1718} & 62.1 & 61.7 & 77.4 & 100\% & 46.2 & 49.1 & 52.3 & 75.4 & \underline{10.5} & 43.8 & 100\% & 100\% \\

\textbf{Hard Routing}~\cite{luo2025mono-internvl} & & & & & & & & & & & & & & & & & & & & \\
\quad \textit{t32-v32}$^\dagger$ & 1. & 30.2 & 29.2 & 58.2 & \textbf{85.6} & 63.3 & 56.3 & 1555 & 59.9 & 57.6 & 76.2 & \textcolor{darkred}{-3.9\%} & 37.4 & 42.4 & 40.4 & 60.7 & 2.5 & 41.2 & \textcolor{darkred}{-26.2\%} & \textcolor{darkred}{-12.3\%} \\
\quad \textit{t48-v16}$^\dagger$ & 1. & 30.3 & 29.7 & 58.6 & 84.6 & 66.0 & 55.4 & 1657 & 62.5 & 61.5 & 76.9 & \textcolor{darkred}{-1.8\%} & 41.9 & 46.6 & 50.2 & 72.4 & 5.3 & 37.5 & \textcolor{darkred}{-14.5\%} & \textcolor{darkred}{-6.6\%} \\

\textbf{MoIIE}~\cite{wang2025moiie} & & & & & & & & & & & & & & & & & & & & \\
\quad \textit{t16-v16-s32}$^\dagger$ & .504 & 32.1 & 29.9 & 58.4 & 85.2 & 65.1 & 55.9 & 1656 & 62.2 & 60.1 & 76.9 & \textcolor{darkred}{-1.5\%} & 41.4 & 45.6 & 49.6 & 72.2 & 5.8 & 40.9 & \textcolor{darkred}{-13.1\%} & \textcolor{darkred}{-5.8\%} \\
\quad \textit{t24-v24-s16}$^\dagger$ & .752 & 31.9 & 29.8 & 58.5 & 84.6 & 63.0 & 55.9 & 1594 & 61.1 & 59.9 & 76.6 & \textcolor{darkred}{-2.6\%} & 41.2 & 42.8 & 47.9 & 68.2 & 2.1 & 42.8 & \textcolor{darkred}{-20.7\%} & \textcolor{darkred}{-9.3\%} \\
\quad \textit{t32-v16-s16}$^\dagger$ & .800 & 30.4 & 29.7 & 58.5 & 84.8 & 64.7 & 55.8 & 1667 & 62.3 & 61.1 & 76.9 & \textcolor{darkred}{-1.9\%} & 42.3 & 50.0 & 50.9 & 72.2 & 6.7 & 40.2 & \textcolor{darkred}{-9.6\%} & \textcolor{darkred}{-4.8\%} \\

\textbf{SMAR}~\cite{xia2025smar} & & & & & & & & & & & & & & & & & & & & \\
\quad \textit{$d_{\text{KL}}$-[0.5, 1.0]}$^*$ & .543 & 31.2 & \textbf{31.5} & 58.7 & 85.5 & \underline{69.3} & 57.3 & 1714 & \underline{63.6} & 61.7 & 77.3 & \textcolor{darkgreen}{+0.6\%} & 42.1 & 43.1 & 47.3 & 67.3 & 7.4 & 45.2 & \textcolor{darkred}{-11.3\%} & \textcolor{darkred}{-3.9\%} \\
\quad \textit{$d_{\text{KL}}$-[1.5, 2.0]}$^*$ & .648 & \underline{32.9} & 30.2 & 59.1 & 85.2 & 68.3 & \underline{57.6} & \textbf{1730} & 60.9 & 58.9 & 77.3 & \textcolor{darkred}{-0.2\%} & 41.7 & 44.3 & 40.6 & 61.7 & 6.3 & \underline{47.7} & \textcolor{darkred}{-15.3\%} & \textcolor{darkred}{-5.8\%} \\
\quad \textit{$d_{\text{KL}}$-[2.5, 3.0]}$^*$ & .743 & 32.2 & 29.9 & 58.3 & 84.9 & 66.8 & 57.1 & 1678 & 60.6 & 59.1 & 77.1 & \textcolor{darkred}{-1.3\%} & 40.7 & 39.2 & 32.3 & 47.5 & 5.5 & 46.5 & \textcolor{darkred}{-25.0\%} & \textcolor{darkred}{-10.2\%} \\

\textbf{\method{}} (ours) & & & & & & & & & & & & & & & & & & & & \\
\quad \textit{attention-soft} & .487 & \textbf{34.7} & \underline{31.0} & \underline{59.6} & 85.1 & 69.0 & \textbf{58.3} & 1706 & 63.5 & \underline{62.3} & \underline{77.4} & \textcolor{darkgreen}{\textbf{+1.8\%}} & \textbf{46.5} & \textbf{56.4} & \textbf{56.0} & \textbf{79.0} & \textbf{10.9} & 46.6 & \textcolor{darkgreen}{\textbf{+6.2\%}} & \textcolor{darkgreen}{\textbf{+3.5\%}} \\
\quad \textit{gaussian-soft} & .440 & 32.4 & 30.8 & \textbf{59.9} & \underline{85.6} & \textbf{69.6} & 57.5 & 1689 & \textbf{65.1} & \textbf{62.5} & \textbf{77.5} & \textcolor{darkgreen}{\underline{+1.3\%}} & \underline{46.4} & \underline{53.6} & \underline{53.9} & \underline{78.0} & 10.2 & \textbf{49.0} & \textcolor{darkgreen}{\underline{+4.2\%}} & \textcolor{darkgreen}{\underline{+2.4\%}} \\
\midrule

\multicolumn{21}{c}{\textit{VLM based on OLMoE (A1B/7B, top-8/64 experts)}} \\
\midrule

\textbf{No Specialization} & .205 & 30.0 & 29.4 & 58.0 & 84.9 & 66.8 & \textbf{56.6} & \textbf{1667} & \textbf{62.4} & 49.4 & 75.7 & 100\% & 49.9 & 48.3 & \underline{59.0} & 79.1 & \textbf{32.6} & 44.2 & 100\% & 100\% \\

\textbf{Hard Routing}~\cite{luo2025mono-internvl} & & & & & & & & & & & & & & & & & & & & \\
\quad \textit{t32-v32}$^\dagger$ & 1. & 28.4 & 27.4 & 57.2 & 84.8 & 61.2 & 50.2 & 1595 & 57.3 & 42.5 & 74.6 & \textcolor{darkred}{-6.1\%} & 36.3 & 35.4 & 40.1 & 59.6 & 3.5 & 42.1 & \textcolor{darkred}{-34.1\%} & \textcolor{darkred}{-16.6\%} \\
\quad \textit{t48-v16}$^\dagger$ & 1. & 31.3 & 29.0 & \underline{58.2} & \underline{85.6} & 65.3 & 53.3 & 1603 & 59.7 & 45.5 & 75.6 & \textcolor{darkred}{-2.0\%} & 43.7 & 42.4 & 51.4 & 73.8 & 13.4 & 47.0 & \textcolor{darkred}{-16.2\%} & \textcolor{darkred}{-7.3\%} \\

\textbf{MoIIE}~\cite{wang2025moiie} & & & & & & & & & & & & & & & & & & & & \\
\quad \textit{t16-v16-s32}$^\dagger$ & .509 & 28.9 & 28.9 & 57.6 & \textbf{85.7} & 65.5 & 53.6 & 1542 & 59.4 & 47.2 & 75.1 & \textcolor{darkred}{-3.0\%} & 43.6 & 44.7 & 51.7 & 74.1 & 14.1 & 47.5 & \textcolor{darkred}{-14.7\%} & \textcolor{darkred}{-7.4\%} \\
\quad \textit{t24-v24-s16}$^\dagger$ & .754 & 31.0 & 28.8 & 57.6 & 85.2 & 62.7 & 52.3 & 1507 & 57.5 & 44.2 & 75.0 & \textcolor{darkred}{-4.2\%} & 41.4 & 44.9 & 45.8 & 71.1 & 3.8 & 46.2 & \textcolor{darkred}{-23.4\%} & \textcolor{darkred}{-11.4\%} \\
\quad \textit{t32-v16-s16}$^\dagger$ & .800 & 32.1 & 29.3 & 57.4 & 85.0 & 63.8 & 53.0 & 1586 & 60.2 & 47.5 & 75.2 & \textcolor{darkred}{-1.8\%} & 41.8 & 37.2 & 49.6 & 74.7 & 13.9 & 46.8 & \textcolor{darkred}{-18.7\%} & \textcolor{darkred}{-8.2\%} \\

\textbf{SMAR}~\cite{xia2025smar} & & & & & & & & & & & & & & & & & & & & \\
\quad \textit{$d_{\text{KL}}$-[0.5, 1.0]}$^*$ & .381 & \underline{32.9} & 29.6 & 58.0 & 84.8 & \underline{67.6} & 54.5 & 1640 & 60.7 & 46.3 & \underline{75.8} & \textcolor{darkred}{-0.3\%} & 48.1 & 50.8 & 58.5 & 77.8 & 25.0 & 43.1 & \textcolor{darkred}{-4.5\%} & \textcolor{darkred}{-1.9\%} \\
\quad \textit{$d_{\text{KL}}$-[1.5, 2.0]}$^*$ & .485 & \textbf{33.1} & 29.6 & 57.8 & 84.8 & 65.5 & 54.0 & 1620 & 61.5 & 47.7 & 75.7 & \textcolor{darkred}{-0.4\%} & 49.4 & 55.1 & 58.7 & 80.2 & 26.5 & 46.0 & \textcolor{darkred}{-0.1\%} & \textcolor{darkred}{-0.3\%} \\
\quad \textit{$d_{\text{KL}}$-[2.5, 3.0]}$^*$ & .645 & 32.4 & \textbf{29.8} & 58.0 & 84.3 & 66.2 & 55.6 & \underline{1650} & 60.6 & 47.8 & 75.8 & \textcolor{darkred}{-0.1\%} & 48.0 & 50.5 & 56.5 & 79.6 & 26.6 & \textbf{50.9} & \textcolor{darkred}{-1.1\%} & \textcolor{darkred}{-0.5\%} \\

\textbf{\method{}} (ours) & & & & & & & & & & & & & & & & & & & & \\
\quad \textit{attention-soft} & .620 & 31.5 & \underline{29.7} & \textbf{58.3} & 84.7 & 66.5 & \underline{55.8} & 1644 & \underline{62.3} & \underline{50.5} & \textbf{75.9} & \textcolor{darkgreen}{\underline{+0.5\%}} & \textbf{50.8} & \textbf{62.2} & \textbf{60.3} & \underline{80.8} & \underline{31.8} & 47.7 & \textcolor{darkgreen}{\textbf{+6.7\%}} & \textcolor{darkgreen}{\textbf{+2.9\%}} \\
\quad \textit{gaussian-soft} & .754 & 31.4 & 29.6 & 58.1 & 85.1 & \textbf{67.9} & 55.4 & 1643 & 62.2 & \textbf{50.6} & 75.7 & \textcolor{darkgreen}{\textbf{+0.6\%}} & \underline{50.1} & \underline{55.6} & 58.7 & \textbf{81.1} & 31.4 & \underline{49.5} & \textcolor{darkgreen}{\underline{+4.3\%}} & \textcolor{darkgreen}{\underline{+2.0\%}} \\
\bottomrule

\end{tabular}
\vspace{-1.0em}
\end{table*}

\subsection{Training Objective and Implementation Details}

In EP deployments, balanced load within each device is essential. We adopt bin-level load-balancing loss:
\begin{equation}
\mathcal{L}_{\text{bal}} = \sum_l \sum_{k=1}^{N_{\text{bins}}} N_B \sum_{e \in \mathbf{B}_k} f_e \, P_e
\end{equation}
where $f_e$ and $P_e$ are computed for tokens in bin $\mathbf{B}_k$ (as defined in \cref{sec:preliminaries}).

The full objective combines task loss, per-bin load-balancing, and inter-bin MI:
\begin{equation}
\mathcal{L} = \mathcal{L}_{\text{task}} + \alpha_{\text{bal}} \mathcal{L}_{\text{bal}} + \alpha_{\text{MI}} \mathcal{L}_{\text{MI}}
\end{equation}
where $\mathcal{L}_{\text{task}}$ is language modeling loss.

We train on 8 NVIDIA A800 GPUs with $N_{\text{bins}} = 8$.
For Gaussian-soft modality scores, the temperature is $\tau = 0.5 \cdot D$.
EMA decay is $\beta = 0.99$ for both Gaussian updates and momentum-adaptive binning.
Loss weights are $\alpha_{\text{bal}} = 0.001$ and $\alpha_{\text{MI}} = 0.0001$.
Further implementation details and ablation studies are in supplementary material.

\section{Results}
\label{sec:results}

We conduct extensive experiments to evaluate \method{} across multiple dimensions: effectiveness on diverse VLM and language-only tasks, ablation studies to validate each design choice, visualization of modality specialization, and computational efficiency under expert-parallel deployment.

\subsection{Experimental Setup}

\noindent\textbf{Model Setup.}
We build VLMs with CLIP ViT-L/14 vision encoder~\cite{radford2021clip}, a 2-layer MLP projector, and MoE language backbones.
We evaluate on four architectures spanning different scales and designs: DeepSeekMoE~\cite{dai2024deepseekmoe}, OL-MoE~\cite{muennighoff2025olmoe}, Moonlight-MoE~\cite{liu2025moonlight-moe}, and Qwen3-MoE~\cite{yang2025qwen3}.
Models are initialized from public checkpoints and fine-tuned with the same protocol.
We compare against four routing strategies: Soft Routing~\cite{mustafa2022limoe}, Hard Routing~\cite{luo2025mono-internvl}, MoIIE~\cite{wang2025moiie} (hybrid-routing), and SMAR~\cite{xia2025smar} (KL-divergence).

\noindent\textbf{Training Data and Benchmarks.}
We follow LLaVA's two-stage training protocol using LLaVA-Pretrain-558K and LLaVA-Instruct-665K datasets respectively~\cite{liu2024llava-1.5}.
Evaluation covers 10 multimodal benchmarks: MMMU-val/test~\cite{yue2024mmmu}, GQA~\cite{hudson2019gqa}, POPE~\cite{li2023pope}, SQA-img~\cite{lu2022science-qa}, TextVQA~\cite{singh2019text-vqa}, MME~\cite{fu2023mme}, MMBench/MMBench-CN~\cite{liu2024mmbench}, VQA-v2~\cite{goyal2017vqa-v2}; and 6 language benchmarks: MMLU~\cite{hendrycks2021mmlu}, HellaSwag~\cite{zellers2019hellaswag}, ARC-C/E~\cite{clark2018arc}, GSM8k~\cite{cobbe2021gsm8k}, TruthfulQA~\cite{lin2021truthfulqa}.

\noindent\textbf{Modality Specialization Index (MSI).}
We introduce MSI to quantify modality-based expert specialization.
Modality affiliation probability for expert $e$ at layer $l$:
\begin{equation}
\tilde{\mathbf{C}}^{(l)}_{m,e} = \frac{\mathbf{C}^{(l)}_{m,e} / \sum_{e'} \mathbf{C}^{(l)}_{m,e'}}{\sum_{m'} \left( \mathbf{C}^{(l)}_{m',e} / \sum_{e'} \mathbf{C}^{(l)}_{m',e'} \right)}
\end{equation}
MSI measures average deviation from uniform distribution across experts and layers:
\begin{equation}
\text{MSI} = \frac{1}{L} \sum_{l=1}^{L} \frac{1}{E} \sum_{e=1}^{E} 2 \cdot \left| \tilde{\mathbf{C}}^{(l)}_{\text{text},e} - 0.5 \right|
\end{equation}
where MSI $\in [0, 1]$: 0 = no specialization, 1 = perfect specialization.

\subsection{Main Results}

We present comprehensive results across all four backbones and 16 benchmarks.
Results on DeepSeekMoE and OLMoE are shown in \cref{tab:main-results-1}.
Results on Moonlight-MoE and Qwen3-MoE are shown in supplementary material.

Across four backbones, \method{} improves over soft routing baseline by 2.2\% average (0.9\% multimodal, 4.2\% language-only).
Soft routing lacks modality-specialization guidance and suffers from data imbalance and load-balancing constraints that constrain expert capability.

Hand-crafted modality specialization (hard/hybrid routing) appears intuitive but inevitably mismatches the complex expert capacity and dynamic data distributions across layers.
Hard routing achieves near-perfect modality specialization but incurs substantial performance degradation (-4.4\% multimodal, -22.2\% language-only).
Hybrid routing (MoIIE) provides some recovery (-3.1\% multimodal, -17.4\% language-only) but remains well below the soft routing baseline.
This reveals that rigid specialization cannot be blindly enforced; it must be learned and adaptively tailored to actual expert capacity and data dynamics.

Why do mainstream MoE-VLMs avoid explicit modality control?
Our hard/hybrid routing results explain this: incorrectly-designed modality specialization schemes incur severe performance penalties.
SMAR introduces an automatic approach through KL divergence regularization rather than explicit hand-crafted assignment.
Since the original SMAR was designed for models with small expert counts (no more than eight), while ours has numerous small experts, we test various KL divergence regularization strengths in our setting.
SMAR achieves significant MSI gain while maintaining more competitive performance: incurring only 1.0\% and 15.1\% performance loss on multimodal and language-only tasks.
However, SMAR suffers from incompatibility between KL divergence regularization and load-balancing constraints (it disabled load-balancing in the final model), making it difficult to leverage specialization in MoE models with numerous small experts.

Our \method{} addresses this by unifying soft modality scores and inter-bin MI objectives, achieving strong modality specialization while simultaneously maximizing expert capacity utilization and maintaining load balance.
Our attention-based soft modality estimation achieves average gains of 1.0\% on multimodal tasks and 4.4\% on language-only tasks, while our Gaussian-statistics variant achieves 0.9\% and 4.1\% respectively.
Both significantly outperform all baseline methods, validating that well-designed learnable specialization schemes can unlock performance gains in large-scale MoE-VLMs while preserving load balance.

\subsection{Ablation Studies}

\subsubsection{Modality Score and Specialization Objective}
\label{sec:ablation-score}

We first demonstrate the effectiveness of soft modality scores.
Hard-score uses binary labels based on input source modality; although it achieves high MSI, it cannot improve model performance.
Both attention-soft and gaussian-soft scores significantly outperform hard-score (\cref{tab:ablation-score}).
Notably, hard-score with MI-based specialization still exceeds the best SMAR configuration, validating the importance of our specialization objective.

\begin{table}
\centering
\caption{
    Ablation on the modality score type (DeepSeekMoE).
}
\label{tab:ablation-score}
\vspace{-0.5em}
\footnotesize
\begin{tabular}{l|c|cc|c}
    \toprule
    \textbf{Method} & \textbf{MSI} & \textbf{Multi-Modal} & \textbf{Language} & \textbf{Overall} \\
    \midrule
    No Specialization & .177 & 100\% & 100\% & 100\% \\
    SMAR (best) & .543 & +0.6\% & -11.3\% & -3.9\% \\
    \midrule
    \textbf{SMoES} & & & & \\
    \quad hard-score & .904 & -0.8\% & +0.5\% & -0.3\% \\
    \quad attention-soft & .487 & \textbf{+1.8\%} & \textbf{+6.2\%} & \textbf{+3.5\%} \\
    \quad gaussian-soft & .440 & \underline{+1.3\%} & \underline{+4.2\%} & \underline{+2.4\%} \\
    \bottomrule
\end{tabular}
\vspace{-0.5em}
\end{table}

We next evaluate our MI-based specialization with expert binning (\cref{tab:ablation-objective}).
Expert-binning itself improves performance (+1.7\%) by providing structure for modality-aware specialization.
Our inter-bin MI objective further boosts gains (+3.5\% for attention-soft, +2.4\% for gaussian-soft), demonstrating that MI effectively guides experts toward coherent modality-specific patterns.
In contrast, KL-divergence in SMAR fails to improve performance.

\begin{table}
\centering
\caption{
    Ablation on inter-bin specialization (DeepSeekMoE).
}
\label{tab:ablation-objective}
\vspace{-0.5em}
\footnotesize
\begin{tabular}{l|c|cc|c}
    \toprule
    \textbf{Method} & \textbf{MSI} & \textbf{Multi-Modal} & \textbf{Language} & \textbf{Overall} \\
    \midrule
    No Specialization & .177 & 100\% & 100\% & 100\% \\
    \quad w/ binning & .415 & +0.9\% & +3.0\% & +1.7\% \\
    \midrule
    w/ inter-bin & & & & \\
    \quad KL & .724 & -1.5\% & -8.5\% & -4.1\% \\
    \quad MI-attention & .487 & \textbf{+1.8\%} & \textbf{+6.2\%} & \textbf{+3.5\%} \\
    \quad MI-gaussian & .440 & \underline{+1.3\%} & \underline{+4.2\%} & \underline{+2.4\%} \\
    \bottomrule
\end{tabular}
\vspace{-1.5em}
\end{table}

\subsubsection{Expert Binning Strategy and Granularity}
\label{sec:ablation-binning}

We evaluate our adaptive binning strategy (based on modality-aware EMA statistics) versus fixed binning (original expert order).
Table \cref{tab:ablation-binning} shows adaptive binning consistently outperforms fixed binning across both binning-only and MI specialization settings, confirming the effectiveness of modality-aware bin formation.

\begin{table}
\centering
\caption{
    Ablation on expert binning strategies (DeepSeekMoE).
}
\label{tab:ablation-binning}
\vspace{-0.5em}
\footnotesize
\begin{tabular}{l|c|cc|c}
    \toprule
    \textbf{Method} & \textbf{MSI} & \textbf{Multi-Modal} & \textbf{Language} & \textbf{Overall} \\
    \midrule
    No Specialization & .177 & 100\% & 100\% & 100\% \\
    w/ binning & & & & \\
    \quad fixed & .357 & +0.9\% & +2.9\% & +1.6\% \\
    \quad adaptive & .415 & +0.9\% & +3.0\% & +1.7\% \\
    \midrule
    attention-soft & & & & \\
    \quad fixed & .450 & +2.0\% & +0.2\% & +1.3\% \\
    \quad adaptive & .487 & +1.8\% & +6.2\% & +3.5\% \\
    \midrule
    gaussian-soft & & & & \\
    \quad fixed & .398 & +1.9\% & -1.0\% & +0.8\% \\
    \quad adaptive & .440 & +1.3\% & +4.2\% & +2.4\% \\
    \bottomrule
\end{tabular}
\end{table}

\begin{figure}[h]
\centering
\includegraphics[width=0.9\columnwidth]{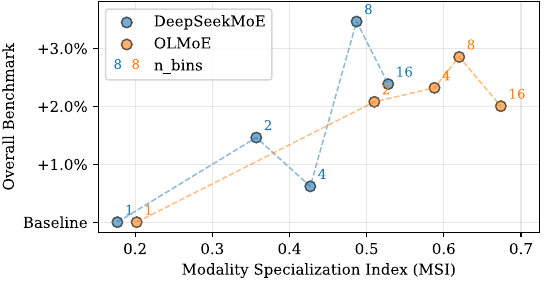}
\vspace{-0.3em}
\caption{
    Ablation on number of expert bins ($\text{SMoES}_{\text{attention-soft}}$).
}
\label{fig:ablation-num-bins}
\vspace{-1.2em}
\end{figure}

We next examine bin granularity by varying the number of bins (\cref{fig:ablation-num-bins}).
Too many bins can increase deployment imbalance despite enabling finer specialization, while too few bins reduce specialization effectiveness. 

\subsection{Visualization and Analysis}

\begin{figure}[ht]
\centering
\vspace{-0.5em}
\includegraphics[width=0.99\columnwidth]{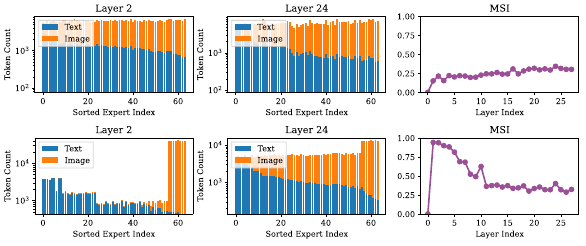}
\vspace{-0.5em}
\caption{
    Routing distribution of tokens to experts in DeepSeekMoE.
    First row: soft baseline; second row: $\text{SMoES}_{\text{attention-soft}}$.
}
\label{fig:routing-dist}
\vspace{-0.8em}
\end{figure}

\method{} achieves clear bin-modality correspondences, as visualized in \cref{fig:routing-dist}.
Early layers exhibit higher MSI and sharper expert-modality separation, while deeper layers show balanced distributions with more modality fusion.

\begin{figure}[t]
\centering
\includegraphics[width=0.99\columnwidth]{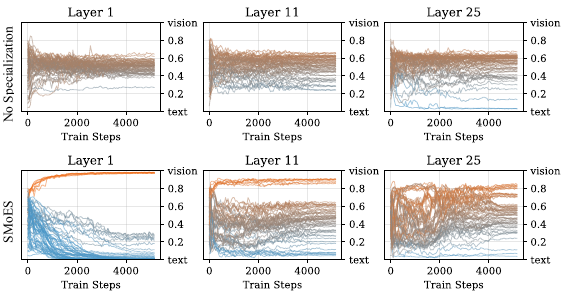}
\vspace{-0.8em}
\caption{
    Evolution of expert specialization during training.
}
\label{fig:training-specialization}
\vspace{-0.8em}
\end{figure}

Expert modality specialization evolves dynamically during training, as shown in \cref{fig:training-specialization}.
Soft routing without specialization results in experts handling both modalities simultaneously, constraining capacity.
Our method shows sharper differentiation in shallow layers where tokens maintain clearer modality identity, and more fusion-aware adaptation in deeper layers.

\begin{figure}[t]
\centering
\includegraphics[width=0.9\columnwidth]{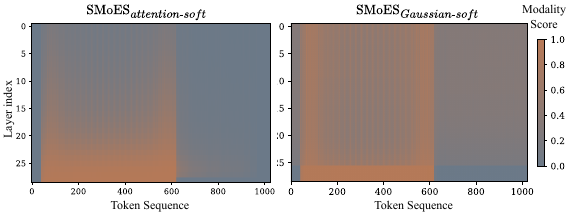}
\vspace{-0.5em}
    \caption{
        Soft modality score across layers in DeepSeekMoE.
    }
    \label{fig:modality-score}
\vspace{-1.5em}
\end{figure}

Our soft modality scores exhibit smooth fusion transitions across layers, as illustrated in \cref{fig:modality-score}.
Both attention-soft and Gaussian-soft estimators show increasing fusion intensity toward deeper layers.
A slight difference is that attention-soft is biased toward text, while Gaussian-soft is biased toward vision in high-fusion regimes.

\subsection{Efficiency Analysis}

\begin{figure}[ht]
\centering
\includegraphics[width=0.99\columnwidth]{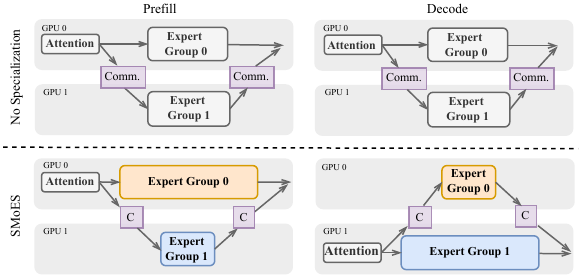}
\vspace{-0.3em}
\caption{
    Expert-parallel (EP) deployment on two GPUs.
}
\label{fig:efficiency-ep}
\vspace{-0.8em}
\end{figure}

Beyond task accuracy, \method{} reduces expert-parallel (EP) communication overhead by aligning expert placement with modality preferences.
We deploy on two NVIDIA Orin GPUs via 10Gb Ethernet using EP, representing a typical edge-side scenario in autonomous vehicles (\cref{fig:efficiency-ep}).
EP avoids weight and KV Cache redundancy compared to TP/DP, making it memory-efficient for edge resources.
Baseline uses synchronous transmission since balanced experts provide no benefit from asynchrony, while \method{}'s increased local expert concentration enables asynchronous transmission, overlapping communication and computation similar to PD separation.

\cref{tab:deployment-communication} shows cross-GPU EP transfer ratios for vision and text tokens at prefill and decode stages on OLMoE.
Expert specialization routes tokens to local experts more frequently, reducing inter-device communication.
We report separate ratios for vision and text tokens (quantities differ between phases) and the overall V+T ratio.
DeDup (de-duplication) is employed to avoid duplicate transmission of tokens routed to experts on the same device in top-$k$ routing.

\begin{table}[t]
    \centering
    \footnotesize
    \setlength{\tabcolsep}{3pt}
    \caption{Cross-GPU EP transfer ratio at prefill/decode stages. V: Vision tokens; T: Text tokens.}
    \label{tab:deployment-communication}
    \vspace{-0.7em}
    \begin{tabular}{lc|ccc|c}
    \toprule
    \multirow{2}{*}{\textbf{Benchmark}} & \multirow{2}{*}{\textbf{Method}} & \multicolumn{3}{c|}{\textbf{Prefill (P)}} & \textbf{Decode (D)} \\
    & & \textbf{V} & \textbf{T} & \textbf{V+T} & \textbf{T} \\
    \midrule
    MMMU & baseline & 97.7\% & 99.5\% & 98.0\% & 86.5\% \\
    (PV:PT:DT=32:8:1) & SMoES & 15.0\% & 99.3\% & 31.1\% & 43.3\% \\
    & $\Delta$ & \textcolor{darkgreen}{↓84.6\%} & \textcolor{darkgreen}{↓0.3\%} & \textcolor{darkgreen}{↓68.3\%} & \textcolor{darkgreen}{↓49.9\%} \\
    \midrule
    SQA-IMG & baseline & 97.5\% & 99.7\% & 98.2\% & 94.9\% \\
    (PV:PT:DT=14:7:1) & SMoES & 13.0\% & 99.2\% & 40.6\% & 49.9\% \\
    & $\Delta$ & \textcolor{darkgreen}{↓86.6\%} & \textcolor{darkgreen}{↓0.5\%} & \textcolor{darkgreen}{↓58.7\%} & \textcolor{darkgreen}{↓47.4\%} \\
    \bottomrule
    \end{tabular}
    \vspace{-0.5em}
\end{table}

\cref{tab:deployment-speed} shows TTFT for Prefill and TPOT for Decode.
Performance improvement stems from reduced communication overhead and parallel execution of computation and communication.
Larger batch sizes in Prefill increase communication proportion, yielding greater gains, while Decode maintains stable improvement ratios across batch sizes due to fewer activated experts.

\begin{table}[t]
    \centering
    \footnotesize
    \setlength{\tabcolsep}{4pt}
    \caption{TTFT and TPOT speed improvement of \method{} compared to soft-routing baseline. $\Delta$: speedup percentage.}
    \label{tab:deployment-speed}
    \vspace{-0.7em}
    \begin{tabular}{lc|cc|cc}
    \toprule
    \multirow{2}{*}{\textbf{Benchmark}} & \multirow{2}{*}{\textbf{Method}} & \multicolumn{2}{c|}{\textbf{Batch Size=1}} & \multicolumn{2}{c}{\textbf{Batch Size=8}} \\
    & & \textbf{TTFT(s)} & \textbf{TPOT(s)} & \textbf{TTFT(s)} & \textbf{TPOT(s)} \\
    \midrule
    \multirow{3}{*}{MMMU} & baseline & 2.810 & 0.786 & 7.949 & 1.414 \\
    & SMoES & \textbf{2.519} & \textbf{0.703} & \textbf{6.203} & \textbf{1.287} \\
    & $\Delta$ & \textcolor{darkgreen}{↓10.3\%} & \textcolor{darkgreen}{↓10.5\%} & \textcolor{darkgreen}{↓22.0\%} & \textcolor{darkgreen}{↓9.0\%} \\
    \midrule
    \multirow{3}{*}{SQA-IMG} & baseline & 1.493 & 0.766 & 5.824 & 1.278 \\
    & SMoES & \textbf{1.356} & \textbf{0.692} & \textbf{4.859} & \textbf{1.134} \\
    & $\Delta$ & \textcolor{darkgreen}{↓9.2\%} & \textcolor{darkgreen}{↓9.7\%} & \textcolor{darkgreen}{↓16.6\%} & \textcolor{darkgreen}{↓11.3\%} \\
    \bottomrule
    \end{tabular}
    \vspace{-1.5em}
\end{table}
\section{Conclusion}
\label{sec:conclusion}

In this paper, we addressed the challenge of modality-guided expert specialization in MoE-VLMs. 
We introduced \method{}, which consists of dynamic soft modality scores that capture layer-dependent fusion patterns, an expert binning mechanism aligned with expert-parallel deployment, and an inter-bin mutual information regularization that encourages coherent modality specialization. 
Extensive experiments across four MoE backbones and 16 benchmarks validate our approach, demonstrating consistent improvements in task accuracy while simultaneously reducing communication overhead and increasing throughput in expert-parallel deployments. 
Our work has some limitations, for example, the Gaussian-soft estimator currently uses a simple unimodal Gaussian with diagonal covariance for efficiency.
Although we have explored GMM as a preliminary extension in the supplementary material, further refining richer density models remains an open problem.
This work demonstrates the value of data-driven modality specialization and opens promising avenues for future exploration in expert specialization for MoE-based multimodal learning.

\clearpage
{
    \small
    \bibliographystyle{ieeenat_fullname}
    \bibliography{main}

@String(AAAI = {AAAI})

@article{team2025kimi-vl,
  title={Kimi-vl technical report},
  author={Team, Kimi and Du, Angang and Yin, Bohong and Xing, Bowei and Qu, Bowen and Wang, Bowen and Chen, Cheng and Zhang, Chenlin and Du, Chenzhuang and Wei, Chu and others},
  journal={arXiv preprint arXiv:2504.07491},
  year={2025}
}

@article{wang2025internvl3.5,
  title={Internvl3. 5: Advancing open-source multimodal models in versatility, reasoning, and efficiency},
  author={Wang, Weiyun and Gao, Zhangwei and Gu, Lixin and Pu, Hengjun and Cui, Long and Wei, Xingguang and Liu, Zhaoyang and Jing, Linglin and Ye, Shenglong and Shao, Jie and others},
  journal={arXiv preprint arXiv:2508.18265},
  year={2025}
}

@article{hong2025glm4.5v,
  title={Glm-4.1 v-thinking: Towards versatile multimodal reasoning with scalable reinforcement learning},
  author={Hong, Wenyi and Yu, Wenmeng and Gu, Xiaotao and Wang, Guo and Gan, Guobing and Tang, Haomiao and Cheng, Jiale and Qi, Ji and Ji, Junhui and Pan, Lihang and others},
  journal={arXiv e-prints},
  pages={arXiv--2507},
  year={2025}
}

@article{shazeer2017moe,
  title={Outrageously large neural networks: The sparsely-gated mixture-of-experts layer},
  author={Shazeer, Noam and Mirhoseini, Azalia and Maziarz, Krzysztof and Davis, Andy and Le, Quoc and Hinton, Geoffrey and Dean, Jeff},
  journal={arXiv preprint arXiv:1701.06538},
  year={2017}
}

@article{lepikhin2020gshard,
  title={Gshard: Scaling giant models with conditional computation and automatic sharding},
  author={Lepikhin, Dmitry and Lee, HyoukJoong and Xu, Yuanzhong and Chen, Dehao and Firat, Orhan and Huang, Yanping and Krikun, Maxim and Shazeer, Noam and Chen, Zhifeng},
  journal={arXiv preprint arXiv:2006.16668},
  year={2020}
}

@article{fedus2022switch-transformers,
  title={Switch transformers: Scaling to trillion parameter models with simple and efficient sparsity},
  author={Fedus, William and Zoph, Barret and Shazeer, Noam},
  journal={Journal of Machine Learning Research},
  volume={23},
  number={120},
  pages={1--39},
  year={2022}
}

@article{kudugunta2021task-moe,
  title={Beyond distillation: Task-level mixture-of-experts for efficient inference},
  author={Kudugunta, Sneha and Huang, Yanping and Bapna, Ankur and Krikun, Maxim and Lepikhin, Dmitry and Luong, Minh-Thang and Firat, Orhan},
  journal={arXiv preprint arXiv:2110.03742},
  year={2021}
}

@article{guo2025deepseek-r1,
  title={Deepseek-r1 incentivizes reasoning in llms through reinforcement learning},
  author={Guo, Daya and Yang, Dejian and Zhang, Haowei and Song, Junxiao and Wang, Peiyi and Zhu, Qihao and Xu, Runxin and Zhang, Ruoyu and Ma, Shirong and Bi, Xiao and others},
  journal={Nature},
  volume={645},
  number={8081},
  pages={633--638},
  year={2025},
  publisher={Nature Publishing Group UK London}
}

@article{wu2024deepseek-vl2,
  title={Deepseek-vl2: Mixture-of-experts vision-language models for advanced multimodal understanding},
  author={Wu, Zhiyu and Chen, Xiaokang and Pan, Zizheng and Liu, Xingchao and Liu, Wen and Dai, Damai and Gao, Huazuo and Ma, Yiyang and Wu, Chengyue and Wang, Bingxuan and others},
  journal={arXiv preprint arXiv:2412.10302},
  year={2024}
}

@article{guo2025seed1.5-vl,
  title={Seed1. 5-vl technical report},
  author={Guo, Dong and Wu, Faming and Zhu, Feida and Leng, Fuxing and Shi, Guang and Chen, Haobin and Fan, Haoqi and Wang, Jian and Jiang, Jianyu and Wang, Jiawei and others},
  journal={arXiv preprint arXiv:2505.07062},
  year={2025}
}

@article{wang2025step-3,
  title={Step-3 is large yet affordable: Model-system co-design for cost-effective decoding},
  author={Wang, Bin and Wang, Bojun and Wan, Changyi and Huang, Guanzhe and Hu, Hanpeng and Jia, Haonan and Nie, Hao and Li, Mingliang and Chen, Nuo and Chen, Siyu and others},
  journal={arXiv preprint arXiv:2507.19427},
  year={2025}
}

@article{bao2022vlmo,
  title={Vlmo: Unified vision-language pre-training with mixture-of-modality-experts},
  author={Bao, Hangbo and Wang, Wenhui and Dong, Li and Liu, Qiang and Mohammed, Owais Khan and Aggarwal, Kriti and Som, Subhojit and Piao, Songhao and Wei, Furu},
  journal={Advances in neural information processing systems},
  volume={35},
  pages={32897--32912},
  year={2022}
}

@inproceedings{wang2023beit3,
  title={Image as a foreign language: Beit pretraining for vision and vision-language tasks},
  author={Wang, Wenhui and Bao, Hangbo and Dong, Li and Bjorck, Johan and Peng, Zhiliang and Liu, Qiang and Aggarwal, Kriti and Mohammed, Owais Khan and Singhal, Saksham and Som, Subhojit and others},
  booktitle={Proceedings of the IEEE/CVF Conference on Computer Vision and Pattern Recognition},
  pages={19175--19186},
  year={2023}
}

@inproceedings{chen2023mod-squad,
  title={Mod-squad: Designing mixtures of experts as modular multi-task learners},
  author={Chen, Zitian and Shen, Yikang and Ding, Mingyu and Chen, Zhenfang and Zhao, Hengshuang and Learned-Miller, Erik G and Gan, Chuang},
  booktitle={Proceedings of the IEEE/CVF Conference on Computer Vision and Pattern Recognition},
  pages={11828--11837},
  year={2023}
}

@article{cai2025ltdr,
  title={Long-Tailed Distribution-Aware Router For Mixture-of-Experts in Large Vision-Language Model},
  author={Cai, Chaoxiang and Yang, Longrong and Chen, Kaibing and Yang, Fan and Li, Xi},
  journal={arXiv preprint arXiv:2507.01351},
  year={2025}
}

@article{yang2024stgc,
  title={Solving token gradient conflict in mixture-of-experts for large vision-language model},
  author={Yang, Longrong and Shen, Dong and Cai, Chaoxiang and Yang, Fan and Gao, Tingting and Zhang, Di and Li, Xi},
  journal={arXiv preprint arXiv:2406.19905},
  year={2024}
}

@article{xia2025smar,
  title={SMAR: Soft Modality-Aware Routing Strategy for MoE-based Multimodal Large Language Models Preserving Language Capabilities},
  author={Xia, Guoyang and Ding, Yifeng and Li, Fengfa and Ren, Lei and Chen, Wei and Feng, Fangxiang and Wang, Xiaojie},
  journal={arXiv preprint arXiv:2506.06406},
  year={2025}
}

@article{mustafa2022limoe,
  title={Multimodal contrastive learning with limoe: the language-image mixture of experts},
  author={Mustafa, Basil and Riquelme, Carlos and Puigcerver, Joan and Jenatton, Rodolphe and Houlsby, Neil},
  journal={Advances in Neural Information Processing Systems},
  volume={35},
  pages={9564--9576},
  year={2022}
}

@inproceedings{chen2024eve,
  title={Eve: Efficient vision-language pre-training with masked prediction and modality-aware moe},
  author={Chen, Junyi and Guo, Longteng and Sun, Jia and Shao, Shuai and Yuan, Zehuan and Lin, Liang and Zhang, Dongyu},
  booktitle={Proceedings of the AAAI Conference on Artificial Intelligence},
  volume={38},
  number={2},
  pages={1110--1119},
  year={2024}
}

@inproceedings{liu2024llava-1.5,
  title={Improved baselines with visual instruction tuning},
  author={Liu, Haotian and Li, Chunyuan and Li, Yuheng and Lee, Yong Jae},
  booktitle={Proceedings of the IEEE/CVF conference on computer vision and pattern recognition},
  pages={26296--26306},
  year={2024}
}

@article{dai2024deepseekmoe,
  title={Deepseekmoe: Towards ultimate expert specialization in mixture-of-experts language models},
  author={Dai, Damai and Deng, Chengqi and Zhao, Chenggang and Xu, RX and Gao, Huazuo and Chen, Deli and Li, Jiashi and Zeng, Wangding and Yu, Xingkai and Wu, Yu and others},
  journal={arXiv preprint arXiv:2401.06066},
  year={2024}
}

@inproceedings{chen2024fast-v,
  title={An image is worth 1/2 tokens after layer 2: Plug-and-play inference acceleration for large vision-language models},
  author={Chen, Liang and Zhao, Haozhe and Liu, Tianyu and Bai, Shuai and Lin, Junyang and Zhou, Chang and Chang, Baobao},
  booktitle={European Conference on Computer Vision},
  pages={19--35},
  year={2024},
  organization={Springer}
}

@inproceedings{jin2024chat-univi,
  title={Chat-univi: Unified visual representation empowers large language models with image and video understanding},
  author={Jin, Peng and Takanobu, Ryuichi and Zhang, Wancai and Cao, Xiaochun and Yuan, Li},
  booktitle={Proceedings of the IEEE/CVF Conference on Computer Vision and Pattern Recognition},
  pages={13700--13710},
  year={2024}
}

@article{meng2024deepstack,
  title={Deepstack: Deeply stacking visual tokens is surprisingly simple and effective for lmms},
  author={Meng, Lingchen and Yang, Jianwei and Tian, Rui and Dai, Xiyang and Wu, Zuxuan and Gao, Jianfeng and Jiang, Yu-Gang},
  journal={Advances in Neural Information Processing Systems},
  volume={37},
  pages={23464--23487},
  year={2024}
}

@article{liu2024deepseek-v3,
  title={Deepseek-v3 technical report},
  author={Liu, Aixin and Feng, Bei and Xue, Bing and Wang, Bingxuan and Wu, Bochao and Lu, Chengda and Zhao, Chenggang and Deng, Chengqi and Zhang, Chenyu and Ruan, Chong and others},
  journal={arXiv preprint arXiv:2412.19437},
  year={2024}
}

@inproceedings{puigcerver2024soft-moe,
  title={From Sparse to Soft Mixtures of Experts},
  author={Joan Puigcerver and Carlos Riquelme Ruiz and Basil Mustafa and Neil Houlsby},
  booktitle={The Twelfth International Conference on Learning Representations},
  year={2024}
}

@article{zhou2022ec-moe,
  title={Mixture-of-experts with expert choice routing},
  author={Zhou, Yanqi and Lei, Tao and Liu, Hanxiao and Du, Nan and Huang, Yanping and Zhao, Vincent and Dai, Andrew M and Le, Quoc V and Laudon, James and others},
  journal={Advances in Neural Information Processing Systems},
  volume={35},
  pages={7103--7114},
  year={2022}
}

@article{chen2023mthl,
  title={An efficient general-purpose modular vision model via multi-task heterogeneous training},
  author={Chen, Zitian and Ding, Mingyu and Shen, Yikang and Zhan, Wei and Tomizuka, Masayoshi and Learned-Miller, Erik and Gan, Chuang},
  journal={arXiv preprint arXiv:2306.17165},
  year={2023}
}

@article{shen2023module-former,
  title={ModuleFormer: Modularity Emerges from Mixture-of-Experts},
  author={Shen, Yikang and Zhang, Zheyu and Cao, Tianyou and Tan, Shawn and Chen, Zhenfang and Gan, Chuang},
  journal={arXiv preprint arXiv:2306.04640},
  year={2023}
}

@article{feng2025co-moe,
  title={CoMoE: Contrastive Representation for Mixture-of-Experts in Parameter-Efficient Fine-tuning},
  author={Feng, Jinyuan and Wei, Chaopeng and Qiu, Tenghai and Hu, Tianyi and Pu, Zhiqiang},
  journal={arXiv preprint arXiv:2505.17553},
  year={2025}
}

@article{zeng2025efficient-moe,
  title={EfficientMoE: Optimizing Mixture-of-Experts Model Training With Adaptive Load Balance},
  author={Zeng, Yan and Huang, Chengchuang and Mei, Yipeng and Zhang, Lifu and Su, Teng and Ye, Wei and Shi, Wenqi and Wang, Shengnan},
  journal={IEEE Transactions on Parallel and Distributed Systems},
  year={2025},
  publisher={IEEE}
}

@article{tang2025moge,
  title={Pangu Pro MoE: Mixture of Grouped Experts for Efficient Sparsity},
  author={Tang, Yehui and Li, Xiaosong and Liu, Fangcheng and Guo, Wei and Zhou, Hang and Wang, Yaoyuan and Han, Kai and Yu, Xianzhi and Li, Jinpeng and Zang, Hui and others},
  journal={arXiv preprint arXiv:2505.21411},
  year={2025}
}

@article{wu2025grove-moe,
  title={Grove moe: Towards efficient and superior moe llms with adjugate experts},
  author={Wu, Haoyuan and Chen, Haoxing and Chen, Xiaodong and Zhou, Zhanchao and Chen, Tieyuan and Zhuang, Yihong and Lu, Guoshan and Huang, Zenan and Zhao, Junbo and Liu, Lin and others},
  journal={arXiv preprint arXiv:2508.07785},
  year={2025}
}

@article{he2025expand-drop,
  title={Capacity-Aware Inference: Mitigating the Straggler Effect in Mixture of Experts},
  author={He, Shwai and Cai, Weilin and Huang, Jiayi and Li, Ang},
  journal={arXiv preprint arXiv:2503.05066},
  year={2025}
}

@article{wang2025aep,
  title={Toward Cost-Efficient Serving of Mixture-of-Experts with Asynchrony},
  author={Wang, Shaoyu and He, Guangrong and Kim, Geon-Woo and Zhou, Yanqi and Park, Seo Jin},
  journal={arXiv preprint arXiv:2505.08944},
  year={2025}
}

@inproceedings{wu2024mole,
  title={Mixture of Lo{RA} Experts},
  author={Xun Wu and Shaohan Huang and Furu Wei},
  booktitle={The Twelfth International Conference on Learning Representations},
  year={2024}
}

@article{chen2024llava-mole,
  title={Llava-mole: Sparse mixture of lora experts for mitigating data conflicts in instruction finetuning mllms},
  author={Chen, Shaoxiang and Jie, Zequn and Ma, Lin},
  journal={arXiv preprint arXiv:2401.16160},
  year={2024}
}

@article{shen2024mome,
  title={Mome: Mixture of multimodal experts for generalist multimodal large language models},
  author={Shen, Leyang and Chen, Gongwei and Shao, Rui and Guan, Weili and Nie, Liqiang},
  journal={Advances in neural information processing systems},
  volume={37},
  pages={42048--42070},
  year={2024}
}

@article{lin2024moe-llave,
  title={Moe-llava: Mixture of experts for large vision-language models},
  author={Lin, Bin and Tang, Zhenyu and Ye, Yang and Cui, Jiaxi and Zhu, Bin and Jin, Peng and Huang, Jinfa and Zhang, Junwu and Pang, Yatian and Ning, Munan and others},
  journal={arXiv preprint arXiv:2401.15947},
  year={2024}
}

@article{zheng2024lte,
  title={Learn to be efficient: Build structured sparsity in large language models},
  author={Zheng, Haizhong and Bai, Xiaoyan and Liu, Xueshen and Mao, Zhuoqing Morley and Chen, Beidi and Lai, Fan and Prakash, Atul},
  journal={Advances in Neural Information Processing Systems},
  volume={37},
  pages={101969--101991},
  year={2024}
}

@article{jing2025evo-moe,
  title={EvoMoE: Expert Evolution in Mixture of Experts for Multimodal Large Language Models},
  author={Jing, Linglin and Gao, Yuting and Wang, Zhigang and Lan, Wang and Tang, Yiwen and Wang, Wenhai and Zhang, Kaipeng and Guo, Qingpei},
  journal={arXiv preprint arXiv:2505.23830},
  year={2025}
}

@inproceedings{shen2023vl-moe,
  title={Scaling Vision-Language Models with Sparse Mixture of Experts},
  author={Sheng Shen and Zhewei Yao and Chunyuan Li and Trevor Darrell and Kurt Keutzer and Yuxiong He},
  booktitle={The 2023 Conference on Empirical Methods in Natural Language Processing},
  year={2023}
}

@article{liang2025mot,
  title={Mixture-of-Transformers: A Sparse and Scalable Architecture for Multi-Modal Foundation Models},
  author={Weixin Liang and LILI YU and Liang Luo and Srini Iyer and Ning Dong and Chunting Zhou and Gargi Ghosh and Mike Lewis and Wen-tau Yih and Luke Zettlemoyer and Xi Victoria Lin},
  journal={Transactions on Machine Learning Research},
  issn={2835-8856},
  year={2025}
}

@article{li2025uni-moe,
  title={Uni-moe: Scaling unified multimodal llms with mixture of experts},
  author={Li, Yunxin and Jiang, Shenyuan and Hu, Baotian and Wang, Longyue and Zhong, Wanqi and Luo, Wenhan and Ma, Lin and Zhang, Min},
  journal={IEEE Transactions on Pattern Analysis and Machine Intelligence},
  year={2025},
  publisher={IEEE}
}

@inproceedings{luo2025mono-internvl,
  title={Mono-internvl: Pushing the boundaries of monolithic multimodal large language models with endogenous visual pre-training},
  author={Luo, Gen and Yang, Xue and Dou, Wenhan and Wang, Zhaokai and Liu, Jiawen and Dai, Jifeng and Qiao, Yu and Zhu, Xizhou},
  booktitle={Proceedings of the Computer Vision and Pattern Recognition Conference},
  pages={24960--24971},
  year={2025}
}

@article{deng2025bagel,
  title={Emerging properties in unified multimodal pretraining},
  author={Deng, Chaorui and Zhu, Deyao and Li, Kunchang and Gou, Chenhui and Li, Feng and Wang, Zeyu and Zhong, Shu and Yu, Weihao and Nie, Xiaonan and Song, Ziang and others},
  journal={arXiv preprint arXiv:2505.14683},
  year={2025}
}

@article{ai2025ming-omni,
  title={Ming-Omni: A Unified Multimodal Model for Perception and Generation},
  author={AI, Inclusion and Gong, Biao and Zou, Cheng and Zheng, Chuanyang and Zhou, Chunluan and Yan, Canxiang and Jin, Chunxiang and Shen, Chunjie and Zheng, Dandan and Wang, Fudong and others},
  journal={arXiv preprint arXiv:2506.09344},
  year={2025}
}

@article{lin2024moma,
  title={Moma: Efficient early-fusion pre-training with mixture of modality-aware experts},
  author={Lin, Xi Victoria and Shrivastava, Akshat and Luo, Liang and Iyer, Srinivasan and Lewis, Mike and Ghosh, Gargi and Zettlemoyer, Luke and Aghajanyan, Armen},
  journal={arXiv preprint arXiv:2407.21770},
  year={2024}
}

@article{wang2025moiie,
  title={Moiie: Mixture of intra-and inter-modality experts for large vision language models},
  author={Wang, Dianyi and Wang, Siyuan and Li, Zejun and Wang, Yikun and Li, Yitong and Tang, Duyu and Shen, Xiaoyu and Huang, Xuanjing and Wei, Zhongyu},
  journal={arXiv preprint arXiv:2508.09779},
  year={2025}
}

@article{welford1962welford,
  title={Note on a method for calculating corrected sums of squares and products},
  author={Welford, Barry Payne},
  journal={Technometrics},
  volume={4},
  number={3},
  pages={419--420},
  year={1962},
  publisher={Taylor \& Francis}
}

@article{dao2022flashattention,
  title={Flashattention: Fast and memory-efficient exact attention with io-awareness},
  author={Dao, Tri and Fu, Dan and Ermon, Stefano and Rudra, Atri and R{\'e}, Christopher},
  journal={Advances in neural information processing systems},
  volume={35},
  pages={16344--16359},
  year={2022}
}

@inproceedings{radford2021clip,
  title={Learning transferable visual models from natural language supervision},
  author={Radford, Alec and Kim, Jong Wook and Hallacy, Chris and Ramesh, Aditya and Goh, Gabriel and Agarwal, Sandhini and Sastry, Girish and Askell, Amanda and Mishkin, Pamela and Clark, Jack and others},
  booktitle={International conference on machine learning},
  pages={8748--8763},
  year={2021},
  organization={PmLR}
}

@inproceedings{muennighoff2025olmoe,
  title={{OLM}oE: Open Mixture-of-Experts Language Models},
  author={Niklas Muennighoff and Luca Soldaini and Dirk Groeneveld and Kyle Lo and Jacob Morrison and Sewon Min and Weijia Shi and Evan Pete Walsh and Oyvind Tafjord and Nathan Lambert and Yuling Gu and Shane Arora and Akshita Bhagia and Dustin Schwenk and David Wadden and Alexander Wettig and Binyuan Hui and Tim Dettmers and Douwe Kiela and Ali Farhadi and Noah A. Smith and Pang Wei Koh and Amanpreet Singh and Hannaneh Hajishirzi},
  booktitle={The Thirteenth International Conference on Learning Representations},
  year={2025}
}

@article{liu2025moonlight-moe,
  title={Muon is scalable for LLM training},
  author={Liu, Jingyuan and Su, Jianlin and Yao, Xingcheng and Jiang, Zhejun and Lai, Guokun and Du, Yulun and Qin, Yidao and Xu, Weixin and Lu, Enzhe and Yan, Junjie and others},
  journal={arXiv preprint arXiv:2502.16982},
  year={2025}
}

@article{yang2025qwen3,
  title={Qwen3 technical report},
  author={Yang, An and Li, Anfeng and Yang, Baosong and Zhang, Beichen and Hui, Binyuan and Zheng, Bo and Yu, Bowen and Gao, Chang and Huang, Chengen and Lv, Chenxu and others},
  journal={arXiv preprint arXiv:2505.09388},
  year={2025}
}

@inproceedings{yue2024mmmu,
  title={Mmmu: A massive multi-discipline multimodal understanding and reasoning benchmark for expert agi},
  author={Yue, Xiang and Ni, Yuansheng and Zhang, Kai and Zheng, Tianyu and Liu, Ruoqi and Zhang, Ge and Stevens, Samuel and Jiang, Dongfu and Ren, Weiming and Sun, Yuxuan and others},
  booktitle={Proceedings of the IEEE/CVF Conference on Computer Vision and Pattern Recognition},
  pages={9556--9567},
  year={2024}
}

@inproceedings{hudson2019gqa,
  title={Gqa: A new dataset for real-world visual reasoning and compositional question answering},
  author={Hudson, Drew A and Manning, Christopher D},
  booktitle={Proceedings of the IEEE/CVF conference on computer vision and pattern recognition},
  pages={6700--6709},
  year={2019}
}

@inproceedings{li2023pope,
  title={Evaluating Object Hallucination in Large Vision-Language Models},
  author={Yifan Li and Yifan Du and Kun Zhou and Jinpeng Wang and Xin Zhao and Ji-Rong Wen},
  booktitle={The 2023 Conference on Empirical Methods in Natural Language Processing},
  year={2023}
}

@article{lu2022science-qa,
  title={Learn to explain: Multimodal reasoning via thought chains for science question answering},
  author={Lu, Pan and Mishra, Swaroop and Xia, Tanglin and Qiu, Liang and Chang, Kai-Wei and Zhu, Song-Chun and Tafjord, Oyvind and Clark, Peter and Kalyan, Ashwin},
  journal={Advances in Neural Information Processing Systems},
  volume={35},
  pages={2507--2521},
  year={2022}
}

@inproceedings{singh2019text-vqa,
  title={Towards vqa models that can read},
  author={Singh, Amanpreet and Natarajan, Vivek and Shah, Meet and Jiang, Yu and Chen, Xinlei and Batra, Dhruv and Parikh, Devi and Rohrbach, Marcus},
  booktitle={Proceedings of the IEEE/CVF conference on computer vision and pattern recognition},
  pages={8317--8326},
  year={2019}
}

@article{fu2023mme,
  title={MME: A Comprehensive Evaluation Benchmark for Multimodal Large Language Models},
  author={Fu, Chaoyou and Chen, Peixian and Shen, Yunhang and Qin, Yulei and Zhang, Mengdan and Lin, Xu and Yang, Jinrui and Zheng, Xiawu and Li, Ke and Sun, Xing and others},
  journal={arXiv preprint arXiv:2306.13394},
  year={2023}
}

@inproceedings{liu2024mmbench,
  title={Mmbench: Is your multi-modal model an all-around player?},
  author={Liu, Yuan and Duan, Haodong and Zhang, Yuanhan and Li, Bo and Zhang, Songyang and Zhao, Wangbo and Yuan, Yike and Wang, Jiaqi and He, Conghui and Liu, Ziwei and others},
  booktitle={European conference on computer vision},
  pages={216--233},
  year={2024},
  organization={Springer}
}

@inproceedings{goyal2017vqa-v2,
  title={Making the v in vqa matter: Elevating the role of image understanding in visual question answering},
  author={Goyal, Yash and Khot, Tejas and Summers-Stay, Douglas and Batra, Dhruv and Parikh, Devi},
  booktitle={Proceedings of the IEEE conference on computer vision and pattern recognition},
  pages={6904--6913},
  year={2017}
}

@inproceedings{hendrycks2021mmlu,
  title={Measuring Massive Multitask Language Understanding},
  author={Dan Hendrycks and Collin Burns and Steven Basart and Andy Zou and Mantas Mazeika and Dawn Song and Jacob Steinhardt},
  booktitle={International Conference on Learning Representations},
  year={2021}
}

@article{zellers2019hellaswag,
  title={Hellaswag: Can a machine really finish your sentence?},
  author={Zellers, Rowan and Holtzman, Ari and Bisk, Yonatan and Farhadi, Ali and Choi, Yejin},
  journal={arXiv preprint arXiv:1905.07830},
  year={2019}
}

@article{clark2018arc,
  title={Think you have solved question answering? try arc, the ai2 reasoning challenge},
  author={Clark, Peter and Cowhey, Isaac and Etzioni, Oren and Khot, Tushar and Sabharwal, Ashish and Schoenick, Carissa and Tafjord, Oyvind},
  journal={arXiv preprint arXiv:1803.05457},
  year={2018}
}

@article{cobbe2021gsm8k,
  title={Training verifiers to solve math word problems},
  author={Cobbe, Karl and Kosaraju, Vineet and Bavarian, Mohammad and Chen, Mark and Jun, Heewoo and Kaiser, Lukasz and Plappert, Matthias and Tworek, Jerry and Hilton, Jacob and Nakano, Reiichiro and others},
  journal={arXiv preprint arXiv:2110.14168},
  year={2021}
}

@article{lin2021truthfulqa,
  title={Truthfulqa: Measuring how models mimic human falsehoods},
  author={Lin, Stephanie and Hilton, Jacob and Evans, Owain},
  journal={arXiv preprint arXiv:2109.07958},
  year={2021}
}
}

\clearpage
\setcounter{page}{1}
\setcounter{section}{0}
\setcounter{table}{0}
\setcounter{figure}{0}
\renewcommand{\thetable}{S\arabic{table}}
\renewcommand{\thefigure}{S\arabic{figure}}
\maketitlesupplementary

\section{Detailed Experimental Setup}

\subsection{Vision Encoder and Projector}

All VLM configurations use the same vision encoder and projector so that backbone and routing changes are the only variables:

\noindent\textbf{Vision Encoder:} CLIP ViT-L/14~\cite{radford2021clip} with 336$\times$336 input resolution, producing 576 visual tokens for every image.
\noindent\textbf{Projector:} A two-layer MLP with GELU activation that maps the 1024-dimensional CLIP features to the hidden size of each MoE backbone (e.g., 2048 for DeepSeekMoE).

\subsection{MoE Backbone Architectures}

We provide detailed specifications for the four MoE-based VLM backbones used in our experiments.
They cover different router designs (softmax vs. sigmoid), expert settings (with or without shared experts), attention variants (MHA, MLA, GQA), and model scales from 7B to 30B parameters.
This variety allows us to validate SMoES under heterogeneous architectures.
\cref{tab:backbone-details} summarizes the main differences.

\noindent\textbf{DeepSeekMoE}~\cite{dai2024deepseekmoe} employs two shared experts together with 64 routed experts under top-6 gating, providing both stable training and sufficient specialization capacity.

\noindent\textbf{OLMoE}~\cite{muennighoff2025olmoe} is an open-source lightweight (only 7B parameters in total) MoE model with 64 experts and top-8 routing. 
Unlike DeepSeekMoE, it does not use shared experts, relying entirely on the routing mechanism for expert selection. 
It is not optimized for multilingual tasks, which can be seen in the MMB-CN benchmark.

\noindent\textbf{Moonlight-MoE}~\cite{liu2025moonlight-moe} is a recent architecture that is similar to DeepSeekMoE in scale. 
However, it differs significantly in its router design: unlike other models that use softmax-based routing, Moonlight-MoE employs sigmoid-based routing with router score bias, which provides a different gating mechanism for expert selection. 
The introduction of router score bias also induces certain fluctuations to the router, for example, hard routing does not perfectly separate visual and text tokens (see MSI in \cref{tab:main-results-2}). 

\noindent\textbf{Qwen3-MoE}~\cite{yang2025qwen3} is a 30B-parameter production model with 128 experts and top-8 routing.
It uses GQA attention and head-wise QK norm, and serves as the strongest baseline among the four backbones.

\subsection{Training Configuration}

We follow the standard LLaVA~\cite{liu2024llava-1.5} two-stage training protocol:

\noindent\textbf{Stage 1: Feature Alignment.} 
We train only the projector while keeping the vision encoder and language model frozen. 
This stage uses 558K image-caption pairs from the LLaVA-Pretrain dataset.

\noindent\textbf{Stage 2: Instruction Tuning.} 
We fine-tune the entire model (except the vision encoder) on the LLaVA-Instruct-665K dataset with diverse vision-language instructions.

We show training details in \cref{tab:training-details}.
All experiments use AdamW optimizer with $\beta_1=0.9$, $\beta_2=0.999$, and weight decay 0.0.
For our method-specific hyperparameters, we set $N_{\text{bins}} = 8$, $\alpha_{\text{bal}} = 0.001$, $\alpha_{\text{MI}} = 0.0001$, EMA decay $\beta = 0.99$, and Gaussian temperature $\tau = 0.5 \cdot D$ as default values across all backbones.

All experiments are conducted on 8 NVIDIA A800-80GB GPUs using mixed-precision training (BF16) and gradient checkpointing to reduce memory usage.
In cases where memory constraints persist, we use gradient accumulation to achieve the effective batch sizes.
During benchmark evaluation, we use greedy search without sampling to produce consistent results.
We set the maximum number of new tokens in generation to 256 for the GSM8k benchmark and 20 for the others.

We do not use dynamic expert migration during training in our experiments. 
However, a reasonable migration frequency according to the expert specialization is encouraged when scaling. 
Thanks to EMA, expert groupings remain stable over a certain period, making the migration overhead reasonable.

\begin{table}[t]
\centering
\caption{Training details.}
\label{tab:training-details}
\small
\begin{tabular}{l|cc}
\toprule
& \textbf{Stage 1} & \textbf{Stage 2} \\
\midrule
Optimizer & AdamW & AdamW \\
$\text{LR}_\text{vision-encoder}$ & 0 & 0 \\
$\text{LR}_\text{projector}$ & 1e-3 & 2e-5 \\
$\text{LR}_\text{language-model}$ & 0 & 2e-5 \\
LR Scheduler & cosine & cosine \\
Warmup Ratio & 0.03 & 0.03 \\
\midrule
Num Samples & 558K & 665K \\
Num Epochs & 1 & 1 \\
Batch Size & 256 & 128 \\
Weight Decay & 0.0 & 0.0 \\
Gradient Clipping & 1.0 & 1.0 \\
\bottomrule
\end{tabular}
\end{table}

\begin{table*}[th]
\centering
\caption{Detailed specifications of MoE backbone architectures.}
\label{tab:backbone-details}
\small
\begin{tabular}{l|cccc}
\toprule
\textbf{Architecture} & \textbf{DeepSeekMoE} & \textbf{OLMoE} & \textbf{Moonlight-MoE} & \textbf{Qwen3-MoE} \\
\midrule
Huggingface Checkpoint & \href{https://huggingface.co/deepseek-ai/deepseek-moe-16b-base}{deepseek-moe-16b-base} & \href{https://huggingface.co/allenai/OLMoE-1B-7B-0125}{OLMoE-1B-7B-0125} & \href{https://huggingface.co/moonshotai/Moonlight-16B-A3B}{Moonlight-16B-A3B} & \href{https://huggingface.co/Qwen/Qwen3-30B-A3B}{Qwen3-30B-A3B} \\
\midrule
Total Parameters & 16B & 7B & 16B & 30B \\
Activated Params & 3B & 1B & 3B & 3B \\
Number of Routed Experts & 64 & 64 & 64 & 128 \\
Top-k Routing & 6 & 8 & 6 & 8 \\
Number of Shared Experts & 2 & 0 & 2 & 0 \\
\midrule
Router Type & softmax & softmax & sigmoid & softmax \\
Router Score Bias & no & no & yes & no \\
\midrule
Hidden Dimension & 2048 & 2048 & 2048 & 2048 \\
Number of Layers & 28 & 16 & 27 & 48 \\
Dense Replace for First k Layers & 1 & 0 & 1 & 0 \\
QK Norm & no & token-wise & no & head-wise \\
Attention Type & MHA & MHA & MLA & GQA \\
\bottomrule
\end{tabular}
\end{table*}

\subsection{Efficient Implementation of SMoES}

Although the attention-accumulated score formulation (\cref{sec:attn-score}) in \method{} uses the attention matrix $\text{Attn}^{(l)}$ for clarity, the aggregation can be efficiently implemented without materializing the full attention matrix, making it compatible with memory-efficient attention kernels such as Flash Attention~\cite{dao2022flashattention}.
Specifically, we concatenate the modality scores $M_{ij,m}^{\text{attn},(l)}$ to the value vectors before applying attention, allowing the kernel to perform the weighted aggregation implicitly.
After the attention operation, we extract the aggregated modality scores from the corresponding dimensions of the attention output.
This implementation strategy preserves the aggregation semantics while leveraging efficient attention implementations without additional memory or computational overhead.

\section{Supplementary Results}

\begin{table}
\centering
\caption{
    Ablation on the modality score type in SMoES.
}
\label{tab:suppl-ablation-score}
\footnotesize
\begin{tabular}{l|c|cc|c}
    \toprule
    \textbf{Method} & \textbf{MSI} & \textbf{Multi-Modal} & \textbf{Language} & \textbf{Overall} \\
    \midrule
    \multicolumn{5}{c}{\textit{VLM based on DeepSeekMoE (A3B/16B, top-6/64 experts)}} \\
    \midrule
    No Specialization & .177 & 100\% & 100\% & 100\% \\
    SMAR (best) & .543 & +0.6\% & -11.3\% & -3.9\% \\
    \midrule
    \textbf{SMoES} & & & & \\
    \quad hard-score & .904 & -0.8\% & +0.5\% & -0.3\% \\
    \quad attention-soft & .487 & \textbf{+1.8\%} & \textbf{+6.2\%} & \textbf{+3.5\%} \\
    \quad gaussian-soft & .440 & \underline{+1.3\%} & \underline{+4.2\%} & \underline{+2.4\%} \\
    \midrule
    \multicolumn{5}{c}{\textit{VLM based on OLMoE (A1B/7B, top-8/64 experts)}} \\
    \midrule
    No Specialization & .205 & 100\% & 100\% & 100\% \\
    SMAR (best) & .485 & -0.4\% & -0.1\% & -0.3\% \\
    \midrule
    \textbf{SMoES} & & & & \\
    \quad hard-score & .756 & -0.0\% & +1.9\% & +0.7\% \\
    \quad attention-soft & .620 & \underline{+0.5\%} & \textbf{+6.7\%} & \textbf{+2.9\%} \\
    \quad gaussian-soft & .754 & \textbf{+0.6\%} & \underline{+4.3\%} & \underline{+2.0\%} \\
    \midrule
    \multicolumn{5}{c}{\textit{VLM based on Moonlight-MoE (A3B/16B, top-6/64 experts)}} \\
    \midrule
    No Specialization & .380 & 100\% & 100\% & 100\% \\
    SMAR (best) & .673 & -2.8\% & -30.3\% & -13.1\% \\
    \midrule
    \textbf{SMoES} & & & & \\
    \quad hard-score & .451 & +0.1\% & +3.1\% & +1.2\% \\
    \quad attention-soft & .442 & \underline{+0.2\%} & \underline{+4.5\%} & \underline{+1.8\%} \\
    \quad gaussian-soft & .449 & \textbf{+0.7\%} & \textbf{+8.0\%} & \textbf{+3.5\%} \\
    \midrule
    \multicolumn{5}{c}{\textit{VLM based on Qwen3-MoE (A3B/30B, top-8/128 experts)}} \\
    \midrule
    No Specialization & .284 & 100\% & \underline{100\%} & 100\% \\
    SMAR (best) & .553 & +0.7\% & -5.4\% & -1.6\% \\
    \midrule
    \textbf{SMoES} & & & & \\
    \quad hard-score & .890 & -1.8\% & -1.5\% & -1.7\% \\
    \quad attention-soft & .726 & \textbf{+1.3\%} & \textbf{+0.1\%} & \textbf{+0.9\%} \\
    \quad gaussian-soft & .566 & \underline{+0.8\%} & -0.2\% & \underline{+0.4\%} \\
    \bottomrule
\end{tabular}
\end{table}

\subsection{Main results on Moonlight-MoE and Qwen3-MoE}

\begin{table*}[t]
    \centering
    \caption{
        Multimodal and language-only results on VLMs based on Moonlight-MoE and Qwen3-MoE.
        \textbf{Bold} and \underline{underline} indicate the first and second best performance.
        MSI denotes Modality Specialization Index. 
        $^\dagger$t/v/s denote the number of text/vision/shared experts.
        $^*$[a, b] denotes KL divergence threshold range.
    }
    \label{tab:main-results-2}
    \footnotesize
    \setlength{\tabcolsep}{2.pt}
    \begin{tabular}{l|c|cccccccccc|c|cccccc|c|c}
    \toprule
    \textbf{Method} & \textbf{MSI} & \multicolumn{11}{c|}{\textbf{Multimodal Tasks (10)}} & \multicolumn{7}{c|}{\textbf{Language-Only Tasks (6)}} & \textbf{Overall} \\
    & & \rotatebox{70}{$\text{MMMU}^{\text{val}}$} & \rotatebox{70}{$\text{MMMU}^{\text{test}}$} & \rotatebox{70}{GQA} & \rotatebox{70}{POPE} & \rotatebox{70}{SQA-IMG} & \rotatebox{70}{TextVQA} & \rotatebox{70}{MME} & \rotatebox{70}{MMB} & \rotatebox{70}{MMB-CN} & \rotatebox{70}{VQAv2} & \rotatebox{70}{\textbf{Avg}} & \rotatebox{70}{MMLU} & \rotatebox{70}{HellaSwag} & \rotatebox{70}{ARC-C} & \rotatebox{70}{ARC-E} & \rotatebox{70}{GSM8k} & \rotatebox{70}{TruthfulQA} & \rotatebox{70}{\textbf{Avg}} & \rotatebox{70}{\textbf{Avg}} \\
    \midrule
    \multicolumn{21}{c}{\textit{VLM based on Moonlight-MoE (A3B/16B, top-6/64 experts)}} \\
    \midrule
    \textbf{No Specialization} & .380 & \underline{34.1} & 31.0 & 58.8 & 85.7 & 67.7 & \underline{58.9} & \textbf{1715} & \underline{67.0} & 63.7 & 77.0 & 100\% & 51.0 & 54.1 & 60.1 & 78.3 & 25.3 & \textbf{51.4} & 100\% & 100\% \\
    
    \textbf{Hard Routing}~\cite{luo2025mono-internvl} & & & & & & & & & & & & & & & & & & & & \\
    \quad \textit{t32-v32}$^\dagger$ & .991 & 27.1 & 27.6 & 57.9 & 85.0 & 59.9 & 54.4 & 1525 & 59.9 & 59.8 & 76.1 & \textcolor{darkred}{-8.2\%} & 33.1 & 39.5 & 40.1 & 53.7 & 3.4 & 43.1 & \textcolor{darkred}{-38.3\%} & \textcolor{darkred}{-19.5\%} \\
    \quad \textit{t48-v16}$^\dagger$ & .952 & 31.2 & 29.6 & 58.5 & 85.4 & 64.1 & 58.1 & 1597 & 64.8 & 60.0 & 76.6 & \textcolor{darkred}{-3.7\%} & 41.8 & 47.3 & 49.5 & 67.9 & 14.1 & 44.4 & \textcolor{darkred}{-19.9\%} & \textcolor{darkred}{-9.8\%} \\
    
    \textbf{MoIIE}~\cite{wang2025moiie} & & & & & & & & & & & & & & & & & & & & \\
    \quad \textit{t16-v16-s32}$^\dagger$ & .686 & 32.2 & 29.9 & 58.4 & \underline{85.7} & 65.5 & 58.2 & 1636 & 63.8 & 61.3 & 76.7 & \textcolor{darkred}{-2.8\%} & 44.1 & 45.6 & 53.2 & 72.9 & 13.1 & 46.3 & \textcolor{darkred}{-17.7\%} & \textcolor{darkred}{-8.3\%} \\
    \quad \textit{t24-v24-s16}$^\dagger$ & .836 & 29.8 & 28.4 & 58.4 & 85.7 & 63.4 & 56.8 & 1615 & 61.7 & 59.8 & 76.5 & \textcolor{darkred}{-5.2\%} & 35.4 & 39.6 & 39.8 & 53.8 & 4.9 & 42.1 & \textcolor{darkred}{-36.9\%} & \textcolor{darkred}{-17.1\%} \\
    \quad \textit{t32-v16-s16}$^\dagger$ & .833 & 32.4 & 29.8 & 58.6 & \textbf{85.7} & 65.1 & 57.4 & 1583 & 63.5 & 60.6 & 76.6 & \textcolor{darkred}{-3.4\%} & 43.7 & 47.5 & 52.8 & 70.3 & 12.4 & 40.3 & \textcolor{darkred}{-20.3\%} & \textcolor{darkred}{-9.7\%} \\
    
    \textbf{SMAR}~\cite{xia2025smar} & & & & & & & & & & & & & & & & & & & & \\
    \quad \textit{$d_{\text{KL}}$-[0.5, 1.0]}$^*$ & .599 & 32.3 & 30.3 & 58.3 & 84.1 & 66.9 & 57.5 & 1590 & 65.7 & 61.3 & 76.7 & \textcolor{darkred}{-2.7\%} & 38.9 & 38.1 & 47.0 & 64.3 & 11.7 & 32.6 & \textcolor{darkred}{-30.6\%} & \textcolor{darkred}{-13.2\%} \\
    \quad \textit{$d_{\text{KL}}$-[1.5, 2.0]}$^*$ & .646 & 31.6 & 29.3 & 58.2 & 84.6 & 66.1 & 55.8 & 1626 & 64.7 & 60.8 & 76.2 & \textcolor{darkred}{-3.7\%} & 36.8 & 39.1 & 39.9 & 55.4 & 6.9 & 26.6 & \textcolor{darkred}{-39.9\%} & \textcolor{darkred}{-17.3\%} \\
    \quad \textit{$d_{\text{KL}}$-[2.5, 3.0]}$^*$ & .673 & 31.9 & 29.6 & 58.1 & 84.9 & \underline{68.1} & 57.7 & \underline{1692} & 63.7 & 59.7 & 76.4 & \textcolor{darkred}{-2.8\%} & 40.0 & 37.2 & 44.6 & 62.8 & 11.1 & 37.6 & \textcolor{darkred}{-30.3\%} & \textcolor{darkred}{-13.1\%} \\
    
    \textbf{\method{}} (ours) & & & & & & & & & & & & & & & & & & & & \\
    \quad \textit{attention-soft} & .442 & 33.7 & \textbf{32.2} & \underline{58.8} & 85.1 & 67.4 & 58.5 & 1668 & \textbf{68.2} & \textbf{64.7} & \underline{77.0} & \textcolor{darkgreen}{\underline{+0.2\%}} & \textbf{52.9} & \underline{55.6} & \underline{67.7} & \underline{84.1} & \underline{26.4} & \underline{49.6} & \textcolor{darkgreen}{\underline{+4.5\%}} & \textcolor{darkgreen}{\underline{+1.8\%}} \\
    \quad \textit{gaussian-soft} & .449 & \textbf{35.7} & \underline{32.0} & \textbf{59.0} & 85.2 & \textbf{68.6} & \textbf{58.9} & 1671 & 67.0 & \underline{64.1} & \textbf{77.0} & \textcolor{darkgreen}{\textbf{+0.7\%}} & \underline{52.9} & \textbf{56.3} & \textbf{68.9} & \textbf{85.0} & \textbf{30.8} & 49.2 & \textcolor{darkgreen}{\textbf{+8.0\%}} & \textcolor{darkgreen}{\textbf{+3.5\%}} \\
    \midrule
    
    \multicolumn{21}{c}{\textit{VLM based on Qwen3-MoE (A3B/30B, top-8/128 experts)}} \\
    \midrule
    \textbf{No Specialization} & .284 & 39.8 & 37.4 & 57.3 & 85.7 & 76.5 & 58.6 & \textbf{1830} & 67.9 & \textbf{71.7} & 75.6 & \textbf{100\%} & \textbf{67.9} & 82.2 & 87.8 & 94.5 & \underline{66.1} & 60.7 & \underline{100\%} & 100\% \\
    
    \textbf{Hard Routing}~\cite{luo2025mono-internvl} & & & & & & & & & & & & & & & & & & & & \\
    \quad \textit{t64-v64}$^\dagger$ & 1. & 37.0 & 31.9 & 56.1 & 85.4 & 66.2 & 55.4 & 1630 & 66.4 & 60.0 & 74.5 & \textcolor{darkred}{-7.4\%} & 51.9 & 68.5 & 75.2 & 87.2 & 34.2 & 54.8 & \textcolor{darkred}{-20.1\%} & \textcolor{darkred}{-12.1\%} \\
    \quad \textit{t96-v32}$^\dagger$ & 1. & 39.6 & 34.6 & 58.0 & \underline{86.6} & 71.7 & 59.1 & 1741 & 68.7 & 66.5 & 75.1 & \textcolor{darkred}{-2.3\%} & 60.5 & 77.9 & 81.3 & 91.1 & 53.3 & 58.8 & \textcolor{darkred}{-8.3\%} & \textcolor{darkred}{-4.5\%} \\
    
    \textbf{MoIIE}~\cite{wang2025moiie} & & & & & & & & & & & & & & & & & & & & \\
    \quad \textit{t32-v32-s64}$^\dagger$ & .509 & 39.9 & 34.6 & 57.0 & 85.9 & 70.8 & 57.7 & 1766 & 69.2 & 66.7 & 75.2 & \textcolor{darkred}{-2.5\%} & 60.9 & 79.3 & 78.7 & 87.3 & 52.6 & 58.2 & \textcolor{darkred}{-9.4\%} & \textcolor{darkred}{-5.1\%} \\
    \quad \textit{t48-v48-s32}$^\dagger$ & .754 & 37.8 & 33.5 & 56.9 & 85.6 & 70.7 & 56.8 & 1649 & 68.7 & 62.0 & 74.9 & \textcolor{darkred}{-5.0\%} & 55.9 & 73.0 & 71.1 & 84.6 & 46.0 & 53.0 & \textcolor{darkred}{-16.9\%} & \textcolor{darkred}{-9.5\%} \\
    \quad \textit{t64-v32-s32}$^\dagger$ & .800 & 38.3 & 34.5 & 57.4 & 85.6 & 72.1 & 58.6 & 1693 & 67.0 & 65.3 & 74.7 & \textcolor{darkred}{-3.6\%} & 60.9 & 78.1 & 83.3 & 91.9 & 54.1 & 59.7 & \textcolor{darkred}{-7.2\%} & \textcolor{darkred}{-5.0\%} \\
    
    \textbf{SMAR}~\cite{xia2025smar} & & & & & & & & & & & & & & & & & & & & \\
    \quad \textit{$d_{\text{KL}}$-[0.5, 1.0]}$^*$ & .747 & \textbf{44.8} & 38.6 & 56.8 & \textbf{86.8} & 72.9 & 57.9 & 1711 & 60.6 & 63.7 & 75.2 & \textcolor{darkred}{-1.9\%} & 56.5 & 69.9 & 75.9 & 86.0 & 56.4 & \textbf{63.4} & \textcolor{darkred}{-10.8\%} & \textcolor{darkred}{-5.2\%} \\
    \quad \textit{$d_{\text{KL}}$-[1.5, 2.0]}$^*$ & .553 & \underline{43.5} & \textbf{40.0} & 56.3 & 85.5 & \textbf{78.2} & 58.8 & 1664 & 69.9 & 69.4 & 75.3 & \textcolor{darkgreen}{+0.7\%} & 60.6 & 74.3 & 82.7 & 92.2 & 61.1 & \underline{63.0} & \textcolor{darkred}{-5.4\%} & \textcolor{darkred}{-1.6\%} \\
    \quad \textit{$d_{\text{KL}}$-[2.5, 3.0]}$^*$ & .647 & 41.9 & 38.7 & 56.2 & 86.1 & \underline{77.6} & \underline{59.8} & 1718 & 69.7 & 69.3 & 75.3 & \textcolor{darkgreen}{+0.4\%} & 60.8 & 74.5 & 81.1 & 91.1 & 60.3 & 59.7 & \textcolor{darkred}{-6.9\%} & \textcolor{darkred}{-2.4\%} \\
    
    \textbf{\method{}} (ours) & & & & & & & & & & & & & & & & & & & & \\
    \quad \textit{attention-soft} & .726 & 40.5 & \underline{38.9} & \underline{58.2} & 85.5 & 75.5 & \textbf{60.8} & \underline{1775} & \textbf{72.7} & \underline{71.0} & \underline{75.9} & \textcolor{darkgreen}{\textbf{+1.3\%}} & \underline{67.7} & \textbf{83.0} & \underline{88.0} & \underline{96.1} & \textbf{66.2} & 59.5 & \textcolor{darkgreen}{\textbf{+0.1\%}} & \textcolor{darkgreen}{\textbf{+0.9\%}} \\
    \quad \textit{gaussian-soft} & .566 & 41.7 & 37.9 & \textbf{58.9} & 85.8 & 76.0 & 59.2 & 1760 & \underline{71.5} & 69.3 & \textbf{76.1} & \textcolor{darkgreen}{\underline{+0.8\%}} & 67.5 & \underline{82.7} & \textbf{88.4} & \textbf{96.7} & 65.2 & 59.0 & \textcolor{darkred}{-0.2\%} & \textcolor{darkgreen}{\underline{+0.4\%}} \\
    \bottomrule
    
    \end{tabular}
    \end{table*}

In addition to the primary evaluation on DeepSeekMoE and OLMoE reported in \cref{tab:main-results-1}, we provide the detailed performance on Moonlight-MoE and Qwen3-MoE in \cref{tab:main-results-2}.
The results demonstrate that SMoES consistently outperforms various routing baselines across these backbones, exhibiting a performance gain trend similar to that observed on DeepSeekMoE and OLMoE.
For Moonlight-MoE, we notice that the improvement in the Modality Specialization Index (MSI) is relatively moderate, increasing to 0.442 from 0.380 in the vanilla soft routing baseline.
This phenomenon likely stems from the architectural characteristics and pretraining strategies of Moonlight-MoE, which appear to inherently foster a certain level of modality differentiation even without explicit routing constraints.
Despite this existing bias, SMoES still effectively refines the routing mechanism to achieve a more structured expert specialization.

Beyond the multimodal performance, we observe a critical limitation in existing modality-specialized methods: they often suffer from substantial performance degradation on language-only tasks.
Such a decline suggests that traditional rigid routing or aggressive specialization constraints can inadvertently compromise the model's fundamental linguistic knowledge.
While the instruction-tuning data incorporates pure text samples, these baseline methods tend to erode the model's ability to handle complex linguistic tasks, particularly those requiring multi-step reasoning like GSM8k.
In the broader context of VLM development, language capabilities are frequently undervalued; however, they remain indispensable for ensuring a high-quality user experience and enabling sophisticated long-context reasoning.
Our results underscore that SMoES successfully maintains robust language performance while enhancing multimodal synergy, which is vital for building truly unified and versatile multimodal models.

\subsection{Gaussian Estimator}

\begin{table}[t]
    \centering
    \small
    \begin{tabular}{l|c|ccc}
        \toprule
        Benchmark & Baseline & k=1 & k=2 & k=4 \\
        \midrule
        multimodal & 100\% & +0.6\% & \textbf{+1.2\%} & +0.8\% \\
        language    & 100\% & +4.3\% & \textbf{+5.6\%} & +4.3\% \\
        Overall     & 100\% & +2.0\% & \textbf{+2.8\%} & +2.1\% \\
        \bottomrule
    \end{tabular}
    \vspace{-5pt}
    \caption{Gaussian Mixture Model (GMM) estimator with different n-components on OLMoE.}
    \label{tab:suppl-gmm-estimator}
\end{table}

In our primary methodology, we employ a unimodal Gaussian Estimator with a diagonal covariance matrix, primarily to ensure computational efficiency and numerical stability in high-dimensional feature spaces.
However, such a simplified distribution may be overly restrictive, potentially failing to capture the intricate, non-linear dependencies and multi-modal structures inherent in fused modality representations.
To investigate the impact of estimator expressiveness, we conduct preliminary experiments exploring more sophisticated density modeling techniques.
While adopting a full covariance matrix could theoretically capture cross-dimension correlations, it introduces $O(D^2)$ parameters and incurs significant overhead for online updates and likelihood computations.
As a more practical alternative, we evaluate the Gaussian Mixture Model (GMM) with $k \in \{2, 4\}$ components, as detailed in \cref{tab:suppl-gmm-estimator}.
Our results indicate that $k=2$ yields measurable performance gains, which aligns with our qualitative observations in \cref{fig:sup-modality-fusion-ds-ol,fig:sup-modality-fusion-ml,fig:sup-modality-fusion-qw} where certain layers exhibit distinct bi-modal or multi-peak token distributions.
These findings suggest that the occasional performance gap between the Gaussian and attention-based estimators likely arises from the limited capacity of a simple Gaussian model to characterize highly complex or heterogeneous token samples.

\begin{table}
    \centering
    \caption{
        Ablation on inter-bin specialization objectives.
        KL: Kullback-Leibler divergence. MI: Mutual Information.
    }
    \label{tab:suppl-ablation-objective}
    \footnotesize
    \begin{tabular}{l|c|cc|c}
        \toprule
        \textbf{Method} & \textbf{MSI} & \textbf{Multi-Modal} & \textbf{Language} & \textbf{Overall} \\
        \midrule
    
        \multicolumn{5}{c}{\textit{VLM based on DeepSeekMoE (A3B/16B, top-6/64 experts)}} \\
        \midrule
        No Specialization & .177 & 100\% & 100\% & 100\% \\
        \quad w/ binning & .415 & +0.9\% & +3.0\% & +1.7\% \\
        \midrule
        w/ Inter-bin & & & & \\
        \quad KL & .724 & -1.5\% & -8.5\% & -4.1\% \\
        \quad MI-attention & .487 & \textbf{+1.8\%} & \textbf{+6.2\%} & \textbf{+3.5\%} \\
        \quad MI-gaussian & .440 & \underline{+1.3\%} & \underline{+4.2\%} & \underline{+2.4\%} \\
        \midrule
    
        \multicolumn{5}{c}{\textit{VLM based on OLMoE (A1B/7B, top-8/64 experts)}} \\
        \midrule
        No Specialization & .205 & 100\% & 100\% & 100\% \\
        \quad w/ binning & .324 & -12.2\% & -9.6\% & -11.2\% \\
        \midrule
        w/ Inter-bin & & & & \\
        \quad KL & .545 & +0.4\% & -0.9\% & -0.1\% \\
        \quad MI-attention & .620 & \underline{+0.5\%} & \textbf{+6.7\%} & \textbf{+2.9\%} \\
        \quad MI-gaussian & .754 & \textbf{+0.6\%} & \underline{+4.3\%} & \underline{+2.0\%} \\
        \midrule
    
        \multicolumn{5}{c}{\textit{VLM based on Moonlight-MoE (A3B/16B, top-6/64 experts)}} \\
        \midrule
        No Specialization & .380 & 100\% & 100\% & 100\% \\
        \quad w/ binning & .441 & -0.0\% & -1.5\% & -0.6\% \\
        \midrule
        w/ Inter-bin & & & & \\
        \quad KL & .748 & -5.9\% & -8.1\% & -6.7\% \\
        \quad MI-attention & .442 & \underline{+0.2\%} & \underline{+4.5\%} & \underline{+1.8\%} \\
        \quad MI-gaussian & .449 & \textbf{+0.7\%} & \textbf{+8.0\%} & \textbf{+3.5\%} \\
        \midrule
    
        \multicolumn{5}{c}{\textit{VLM based on Qwen3-MoE (A3B/30B, top-8/128 experts)}} \\
        \midrule
        No Specialization & .284 & 100\% & 100\% & 100\% \\
        \quad w/ binning & .468 & +0.6\% & -0.1\% & +0.3\% \\
        \midrule
        w/ Inter-bin & & & & \\
        \quad KL & .540 & -2.0\% & \textbf{+1.0\%} & -1.0\% \\
        \quad MI-attention & .726 & \textbf{+1.3\%} & \underline{+0.1\%} & \underline{+0.9\%} \\
        \quad MI-gaussian & .566 & \underline{+0.8\%} & -0.2\% & \underline{+0.4\%} \\
        \bottomrule
    \end{tabular}
    \end{table}

\subsection{Ablation study}

Ablation on the modality score type on all four MoE backbones is shown in \cref{tab:suppl-ablation-score}.
Our attention-soft and gaussian-soft scores outperform the hard routing score (0/1 hard-coded) on all four MoE backbones.
This demonstrates the importance of soft modality scores in modality differentiation.
Furthermore, even the hard modality score can outperform SMAR, indicating that MI-based modality differentiation is more effective than KL-based approaches.

\begin{table}
    \centering
    \caption{
        Ablation on expert binning strategy.
    }
    \label{tab:suppl-ablation-binning}
    \footnotesize
    \begin{tabular}{l|c|cc|c}
        \toprule
        \textbf{Method} & \textbf{MSI} & \textbf{Multi-Modal} & \textbf{Language} & \textbf{Overall} \\
        \midrule
        
        \multicolumn{5}{c}{\textit{VLM based on DeepSeekMoE (A3B/16B, top-6/64 experts)}} \\
        \midrule
        No Specialization & .177 & 100\% & 100\% & 100\% \\
        w/ binning & & & & \\
        \quad fixed & .357 & +0.9\% & +2.9\% & +1.6\% \\
        \quad adaptive & .415 & +0.9\% & +3.0\% & +1.7\% \\
        \midrule
        attention-soft & & & & \\
        \quad fixed & .450 & +2.0\% & +0.2\% & +1.3\% \\
        \quad adaptive & .487 & +1.8\% & +6.2\% & +3.5\% \\
        \midrule
        gaussian-soft & & & & \\
        \quad fixed & .398 & +1.9\% & -1.0\% & +0.8\% \\
        \quad adaptive & .440 & +1.3\% & +4.2\% & +2.4\% \\
        \midrule
    
        \multicolumn{5}{c}{\textit{VLM based on OLMoE (A1B/7B, top-8/64 experts)}} \\
        \midrule
        No Specialization & .205 & 100\% & 100\% & 100\% \\
        w/ binning & & & & \\
        \quad fixed & .324 & -18.6\% & -10.5\% & -15.6\% \\
        \quad adaptive & .324 & -12.2\% & -9.6\% & -11.2\% \\
        \midrule
        attention-soft & & & & \\
        \quad fixed & .355 & +0.1\% & +3.3\% & +1.4\% \\
        \quad adaptive & .620 & +0.5\% & +6.7\% & +2.9\% \\
        \midrule
        gaussian-soft & & & & \\
        \quad fixed & .433 & -0.6\% & +1.8\% & +0.5\% \\
        \quad adaptive & .754 & +0.6\% & +4.3\% & +2.0\% \\
        \midrule
    
        \multicolumn{5}{c}{\textit{VLM based on Moonlight-MoE (A3B/16B, top-6/64 experts)}} \\
        \midrule
        No Specialization & .380 & 100\% & 100\% & 100\% \\
        w/ binning & & & & \\
        \quad fixed & .445 & -9.8\% & -37.3\% & -20.2\% \\
        \quad adaptive & .441 & -0.0\% & -1.5\% & -0.6\% \\
        \midrule
        attention-soft & & & & \\
        \quad fixed & .377 & +0.1\% & -0.6\% & -0.2\% \\
        \quad adaptive & .442 & +0.2\% & +4.5\% & +1.8\% \\
        \midrule
        gaussian-soft & & & & \\
        \quad fixed & .388 & +0.3\% & +5.0\% & +2.1\% \\
        \quad adaptive & .449 & +0.7\% & +8.0\% & +3.5\% \\
        \midrule
    
        \multicolumn{5}{c}{\textit{VLM based on Qwen3-MoE (A3B/30B, top-8/128 experts)}} \\
        \midrule
        No Specialization & .284 & 100\% & 100\% & 100\% \\
        w/ binning & & & & \\
        \quad fixed & .308 & -0.1\% & +0.6\% & +0.1\% \\
        \quad adaptive & .468 & +0.6\% & -0.1\% & +0.3\% \\
        \midrule
        attention-soft & & & & \\
        \quad fixed & .325 & +1.3\% & +1.8\% & +1.5\% \\
        \quad adaptive & .726 & +1.3\% & +0.1\% & +0.9\% \\
        \midrule
        gaussian-soft & & & & \\
        \quad fixed & .332 & +0.6\% & +0.0\% & +0.4\% \\
        \quad adaptive & .566 & +0.8\% & -0.2\% & +0.4\% \\
    
        \bottomrule
    \end{tabular}
    \end{table}

Ablation on inter-bin specialization objective is shown in \cref{tab:suppl-ablation-objective}.
Expert binning can increase MSI since it provides the feasibility for modality differentiation. 
Inter-bin MI-based specialization further enhances modality differentiation and improves model performance.
In contrast, KL-based inter-bin modality differentiation, despite achieving higher modality differentiation degrees on DeepSeekMoE and Moonlight-MoE, degrades model performance, which is unacceptable. 

Ablation on expert binning strategy is shown in \cref{tab:suppl-ablation-binning}.
We compare our momentum-adaptive binning strategy (adaptive) to the static, predefined binning (fixed) strategy.
In all cases, adaptive binning is superior to fixed binning in terms of model performance.
Furthermore, in most cases, adaptive binning yields a higher degree of modality specialization.
This is because the pretrained MoE backbone already exhibits certain expert capability tendencies, and forcing predefined modality bins for experts would interfere with their capability expression.
Furthermore, during training, experts may gradually differentiate, and fixed binning may constrain the experts' modality differentiation capacity. 

Ablation on Gaussian temperature in gaussian-soft SMoES is shown in \cref{tab:suppl-ablation-gaussian-temperature}.
We evaluate $\tau \in \{0.05D, 0.1D, 0.3D, 0.5D, 1.0D\}$, where $D$ is the feature dimension.
As shown in the table, $0.5D$ provides the best balance between confident and smooth modality scores.
The table shows that both excessively high and low temperatures lead to performance degradation.
Moderate temperature levels yield similar and stable performance improvements.
Therefore, we adopt a moderate value of $0.5D$ as the default setting.

\begin{table}
\centering
\caption{
    Ablation on Gaussian temperature.
    $D$: number of the feature dimension.
}
\label{tab:suppl-ablation-gaussian-temperature}
\footnotesize
\begin{tabular}{l|c|cc|c}
    \toprule
    \textbf{Method} & \textbf{MSI} & \textbf{Multi-Modal} & \textbf{Language} & \textbf{Overall} \\
    \midrule
    
    \multicolumn{5}{c}{\textit{VLM based on DeepSeekMoE (A3B/16B, top-6/64 experts)}} \\
    \midrule
    No Specialization & .177 & 100\% & 100\% & 100\% \\
    \midrule
    gaussian-$\tau$ & & & & \\
    \quad $0.05D$ & .516 & -0.9\% & \textbf{+4.8\%} & +1.2\% \\
    \quad $0.1D$ & .574 & +0.6\% & +4.2\% & +1.9\% \\
    \quad $0.3D$ & .564 & \textbf{+1.6\%} & +1.4\% & +1.5\% \\
    \quad \textit{$0.5D$ (default)} & .440 & +1.3\% & \underline{+4.2\%} & \textbf{+2.4\%} \\
    \quad $0.7D$ & .444 & \underline{+1.4\%} & +3.1\% & \underline{+2.1\%} \\
    \quad $1.0D$ & .475 & +0.2\% & +2.4\% & +1.1\% \\
    \midrule

    \multicolumn{5}{c}{\textit{VLM based on OLMoE (A1B/7B, top-8/64 experts)}} \\
    \midrule
    No Specialization & .205 & 100\% & 100\% & 100\% \\
    \midrule
    gaussian-$\tau$ & & & & \\
    \quad $0.05D$ & .744 & -0.4\% & +0.6\% & -0.0\% \\
    \quad $0.1D$ & .744 & -0.1\% & \textbf{+4.3\%} & \underline{+1.7\%} \\
    \quad $0.3D$ & .759 & -0.4\% & \underline{+4.3\%} & +1.4\% \\
    \quad \textit{$0.5D$ (default)} & .754 & \textbf{+0.6\%} & +4.3\% & \textbf{+2.0\%} \\
    \quad $0.7D$ & .745 & +0.2\% & +3.9\% & +1.6\% \\
    \quad $1.0D$ & .740 & \underline{+0.5\%} & -2.0\% & -0.5\% \\

    \bottomrule
\end{tabular}
\end{table}

Ablation on modality specialization loss weight ($\alpha_{\text{MI}}$) and load balance loss weight ($\alpha_{\text{bal}}$) in attention-soft SMoES is shown in \cref{tab:suppl-ablation-alpha}.
We evaluate $\alpha_{\text{MI}}$ and $\alpha_{\text{bal}}$ in $\{1e-1, 1e-2, 1e-3, 1e-4, 1e-5\}$.
The table shows that when $\alpha_{\text{MI}}$ is relatively large, it is possible to achieve substantial model performance improvements, but it may also cause training instability.
Therefore, we choose a relatively low $\alpha_{\text{MI}}$ ($1e-4$) to obtain stable performance improvements.
For $\alpha_{\text{bal}}$, the results may vary across different models.
For example, on DeepSeekMoE, using a smaller balance loss can further improve model performance, while on OLMoE, a moderate balance loss is preferred.
To achieve stable performance and ensure the magnitude is consistent with the balance loss of the soft routing baseline, we choose $1e-3$ as the weight for the balance loss.

\begin{table}
\centering
\caption{
    Ablation on loss alpha ($\text{SMoES}_{\text{attention-soft}}$).
}
\label{tab:suppl-ablation-alpha}
\footnotesize
\begin{tabular}{l|c|cc|c}
    \toprule
    \textbf{Method} & \textbf{MSI} & \textbf{Multi-Modal} & \textbf{Language} & \textbf{Overall} \\
    \midrule
    
    \multicolumn{5}{c}{\textit{VLM based on DeepSeekMoE (A3B/16B, top-6/64 experts)}} \\
    \midrule
    No Specialization & .177 & 100\% & 100\% & 100\% \\
    \midrule
    $\alpha_{\text{MI}}$ & & & & \\
    \quad 1e-1 & .533 & \textbf{+3.0\%} & \textbf{+16.1\%} & \textbf{+8.6\%} \\
    \quad 1e-2 & .716 & -11.9\% & +1.0\% & -6.3\% \\
    \quad 1e-3 & .610 & -1.0\% & +4.9\% & +1.5\% \\
    \quad \textit{1e-4 (default)} & .487 & \underline{+1.8\%} & \underline{+6.2\%} & \underline{+3.5\%} \\
    \quad 1e-5 & .417 & -0.0\% & +4.0\% & +1.7\% \\
    \midrule
    $\alpha_{\text{bal}}$ & & & & \\
    \quad 1e-1 & .290 & -43.2\% & -48.4\% & -45.3\% \\
    \quad 1e-2 & .410 & -7.1\% & -17.8\% & -11.3\% \\
    \quad \textit{1e-3 (default)} & .487 & +1.8\% & \underline{+6.2\%} & +3.5\% \\
    \quad 1e-4 & .578 & \textbf{+3.9\%} & \textbf{+8.1\%} & \textbf{+5.2\%} \\
    \quad 1e-5 & .550 & \underline{+3.9\%} & +6.0\% & \underline{+4.5\%} \\
    \midrule

    \multicolumn{5}{c}{\textit{VLM based on OLMoE (A1B/7B, top-8/64 experts)}} \\
    \midrule
    No Specialization & .205 & 100\% & 100\% & 100\% \\
    \midrule
    $\alpha_{\text{MI}}$ & & & & \\
    \quad 1e-1 & .684 & \underline{+1.8\%} & +4.6\% & +2.8\% \\
    \quad 1e-2 & .871 & \textbf{+2.0\%} & \textbf{+8.3\%} & \textbf{+4.3\%} \\
    \quad 1e-3 & .982 & -0.2\% & +2.2\% & +0.7\% \\
    \quad \textit{1e-4 (default)} & .620 & +0.5\% & \underline{+6.7\%} & \underline{+2.9\%} \\
    \quad 1e-5 & .340 & +0.3\% & +3.9\% & +1.7\% \\

    \midrule
    $\alpha_{\text{bal}}$ & & & & \\
    \quad 1e-1 & .738 & -15.4\% & -27.9\% & -20.1\% \\
    \quad 1e-2 & .796 & -0.3\% & -1.8\% & -0.9\% \\
    \quad \textit{1e-3 (default)} & .620 & \textbf{+0.5\%} & \textbf{+6.7\%} & \textbf{+2.9\%} \\
    \quad 1e-4 & .654 & \underline{+0.4\%} & +4.9\% & +2.1\% \\
    \quad 1e-5 & .702 & +0.3\% & \underline{+6.0\%} & \underline{+2.6\%} \\

    \bottomrule
\end{tabular}
\end{table}

\subsection{Efficiency Analysis}

\begin{table*}[t]
    \centering
    \small
    \caption{
        Cross-GPU EP transfer ratio at prefill and decode stages for OLMoE. 
        V: Vision tokens; T: Text tokens. 
        \textit{(PV : PT : DT)}: token count ratio for prefill-vision, prefill-text and decode-text.
        DeDup: de-duplication of tokens routed to experts on the same device in top-K routing.
    }
    \label{tab:suppl-deployment-communication}
    \begin{tabular}{l|ccc|c|ccc|c}
    \toprule
    \multirow{2}{*}{\textbf{Method}} & \multicolumn{4}{c|}{Top-K w/o DeDup } & \multicolumn{4}{c}{Top-K w/ DeDup} \\
    & \multicolumn{3}{c|}{\textbf{Prefill (P)}} & \textbf{Decode (D)} & \multicolumn{3}{c|}{\textbf{Prefill (P)}} & \textbf{Decode (D)} \\
    & \textbf{V} & \textbf{T} & \textbf{V+T} & \textbf{T} & \textbf{V} & \textbf{T} & \textbf{V+T} & \textbf{T} \\
    \midrule
    
    \multicolumn{9}{c}{\textit{MMMU (PV : PT : DT = 79\% : 19\% : 2\%)}} \\
    \midrule
    Baseline (No Specialization) & 42.7\% & 70.4\% & 48.0\% & 37.4\% & 97.7\% & 99.5\% & 98.0\% & 86.5\% \\
    \addlinespace[0.5em]
    $\text{SMoES}_{\text{attention-soft}}$ & 8.8\% & 71.7\% & 20.7\% & 34.3\% & 33.6\% & 98.9\% & 46.1\% & 88.9\% \\
    & \textcolor{darkgreen}{↓ 79.4\%} & \textcolor{darkred}{↑ 1.0\%} & \textcolor{darkgreen}{↓ 56.9\%} & \textcolor{darkgreen}{↓ 8.3\%} & \textcolor{darkgreen}{↓ 65.6\%} & \textcolor{darkgreen}{↓ 0.6\%} & \textcolor{darkgreen}{↓ 53.0\%} & \textcolor{darkred}{↑ 2.8\%} \\
    \addlinespace[0.3em]
    $\text{SMoES}_{\text{gaussian-soft}}$ & 3.3\% & 76.7\% & 17.3\% & 16.0\% & 15.0\% & 99.3\% & 31.1\% & 43.3\% \\
    & \textcolor{darkgreen}{↓ 92.3\%} & \textcolor{darkred}{↑ 8.9\%} & \textcolor{darkgreen}{↓ 63.9\%} & \textcolor{darkgreen}{↓ 57.3\%} & \textcolor{darkgreen}{↓ 84.6\%} & \textcolor{darkgreen}{↓ 0.3\%} & \textcolor{darkgreen}{↓ 68.3\%} & \textcolor{darkgreen}{↓ 49.9\%} \\
    \midrule
    
    \multicolumn{9}{c}{\textit{SQA-IMG (PV : PT : DT = 65\% : 30\% : 5\%)}} \\
    \midrule
    Baseline (No Specialization) & 42.5\% & 74.1\% & 52.6\% & 49.0\% & 97.5\% & 99.7\% & 98.2\% & 94.9\%\\
    \addlinespace[0.5em]
    $\text{SMoES}_{\text{attention-soft}}$ & 8.4\% & 76.3\% & 30.1\% & 41.7\% & 33.2\% & 99.3\% & 54.3\% & 90.0\% \\
    & \textcolor{darkgreen}{↓80.3\%} & \textcolor{darkred}{↑ 2.9\%} & \textcolor{darkgreen}{↓ 42.9\%} & \textcolor{darkgreen}{↓15.0\%} & \textcolor{darkgreen}{↓ 65.9\%} & \textcolor{darkgreen}{↓ 0.4\%} & \textcolor{darkgreen}{↓ 44.7\%} & \textcolor{darkgreen}{↓ 5.2\%} \\
    \addlinespace[0.3em]
    $\text{SMoES}_{\text{gaussian-soft}}$ & 2.7\% & 80.3\% & 27.5\% & 17.7\% & 13.0\% & 99.2\% & 40.6\% & 49.9\% \\
    & \textcolor{darkgreen}{↓ 93.6\%} & \textcolor{darkred}{↑ 8.2\%} & \textcolor{darkgreen}{↓ 47.7\%} & \textcolor{darkgreen}{↓ 63.9\%} & \textcolor{darkgreen}{↓ 86.6\%} & \textcolor{darkgreen}{↓ 0.5\%} & \textcolor{darkgreen}{↓ 58.7\%} & \textcolor{darkgreen}{↓ 47.4\%} \\
    \midrule
    
    \multicolumn{9}{c}{\textit{POPE (PV : PT : DT = 86\% : 11\% : 3\%)}} \\
    \midrule
    Baseline (No Specialization) & 42.9\% & 85.7\% & 47.8\% & 37.7\% & 97.8\% & 100\% & 98.1\% & 87.8\% \\
    \addlinespace[0.5em]
    $\text{SMoES}_{\text{attention-soft}}$ & 8.0\% & 88.2\% & 17.1\% & 31.4\% & 33.4\% & 99.9\% & 41.0\% & 84.4\% \\
    & \textcolor{darkgreen}{↓ 81.4\%} & \textcolor{darkred}{↑ 2.9\%} & \textcolor{darkgreen}{↓ 64.1\%} & \textcolor{darkgreen}{↓ 16.7\%} & \textcolor{darkgreen}{↓ 65.9\%} & \textcolor{darkgreen}{↓ 0.1\%} & \textcolor{darkgreen}{↓ 58.2\%} & \textcolor{darkgreen}{↓ 3.8\%} \\
    \addlinespace[0.3em]
    $\text{SMoES}_{\text{gaussian-soft}}$ & 1.8\% & 91.5\% & 12.0\% & 14.5\% & 9.0\% & 99.8\% & 19.4\% & 40.8\% \\
    & \textcolor{darkgreen}{↓ 95.8\%} & \textcolor{darkred}{↑ 6.8\%} & \textcolor{darkgreen}{↓ 74.8\%} & \textcolor{darkgreen}{↓ 61.5\%} & \textcolor{darkgreen}{↓ 90.8\%} & \textcolor{darkgreen}{↓ 0.2\%} & \textcolor{darkgreen}{↓ 80.2\%} & \textcolor{darkgreen}{↓ 53.6\%} \\
    \midrule
    
    \multicolumn{9}{c}{\textit{GQA (PV : PT : DT = 86\% : 11\% : 3\%)}} \\
    \midrule
    Baseline (No Specialization) & 43.0\% & 85.5\% & 47.9\% & 38.3\% & 97.9\% & 100\% & 98.1\% & 89.2\% \\
    \addlinespace[0.5em]
    $\text{SMoES}_{\text{attention-soft}}$ & 8.2\% & 87.7\% & 17.4\% & 32.1\% & 34.0\% & 99.9\% & 41.7\% & 84.8\% \\
    & \textcolor{darkgreen}{↓ 80.9\%} & \textcolor{darkred}{↑ 2.6\%} & \textcolor{darkgreen}{↓ 63.6\%} & \textcolor{darkgreen}{↓ 16.2\%} & \textcolor{darkgreen}{↓ 65.2\%} & \textcolor{darkgreen}{↓ 0.1\%} & \textcolor{darkgreen}{↓ 57.5\%} & \textcolor{darkgreen}{↓ 5.0\%} \\
    \addlinespace[0.3em]
    $\text{SMoES}_{\text{gaussian-soft}}$ & 1.9\% & 90.9\% & 12.2\% & 16.2\% & 9.3\% & 99.8\% & 19.8\% & 42.4\% \\
    & \textcolor{darkgreen}{↓ 95.7\%} & \textcolor{darkred}{↑ 6.4\%} & \textcolor{darkgreen}{↓ 74.5\%} & \textcolor{darkgreen}{↓ 57.7\%} & \textcolor{darkgreen}{↓ 90.5\%} & \textcolor{darkgreen}{↓ 0.2\%} & \textcolor{darkgreen}{↓ 79.8\%} & \textcolor{darkgreen}{↓ 52.5\%} \\
    \midrule
    
    \multicolumn{9}{c}{\textit{TextVQA (PV : PT : DT = 80\% : 17\% : 3\%)}} \\
    \midrule
    baseline & 43.6\% & 72.5\% & 48.6\% & 43.1\% & 97.9\% & 99.1\% & 98.1\% & 90.7\% \\
    \addlinespace[0.5em]
    $\text{SMoES}_{\text{attention-soft}}$ & 8.9\% & 72.0\% & 19.9\% & 34.5\% & 35.8\% & 97.1\% & 46.5\% & 82.6\% \\
    & \textcolor{darkgreen}{↓ 79.5\%} & \textcolor{darkgreen}{↓ 0.8\%} & \textcolor{darkgreen}{↓ 59.0\%} & \textcolor{darkgreen}{↓ 19.9\%} & \textcolor{darkgreen}{↓ 63.5\%} & \textcolor{darkgreen}{↓ 2.1\%} & \textcolor{darkgreen}{↓ 52.6\%} & \textcolor{darkgreen}{↓ 8.9\%} \\
    \addlinespace[0.3em]
    $\text{SMoES}_{\text{gaussian-soft}}$ & 2.6\% & 77.2\% & 15.6\% & 21.0\% & 12.6\% & 97.5\% & 27.4\% & 47.1\% \\
    & \textcolor{darkgreen}{↓ 93.9\%} & \textcolor{darkred}{↑ 6.4\%} & \textcolor{darkgreen}{↓ 67.8\%} & \textcolor{darkgreen}{↓ 51.1\%} & \textcolor{darkgreen}{↓ 87.1\%} & \textcolor{darkgreen}{↓ 1.6\%} & \textcolor{darkgreen}{↓ 72.0\%} & \textcolor{darkgreen}{↓ 48.0\%} \\
    \bottomrule
    \end{tabular}
    \end{table*}

To evaluate the practical advantages of our SMoES method in real-world scenarios, we conduct edge-side inference experiments comparing our \method{} against baseline models without modality specialization. 
These experiments demonstrate the communication efficiency gains achieved through expert specialization, particularly in resource-constrained edge deployment settings.

\begin{table*}[t]
    \centering
    \footnotesize
    \setlength{\tabcolsep}{2pt}
    \caption{TTFT (Time to First Token, Prefill) and TPOT (Time Per Output Token, Decode) speed improvement of $\method{}_{\text{attention-soft}}$ compared to soft-routing baseline. $\Delta$: speedup percentage.}
    \label{tab:suppl-deployment-speed}
    \begin{tabular}{lc|cc|cc|cc|cc|cc|cc}
    \toprule
    \multirow{2}{*}{\textbf{Benchmark}} & \multirow{2}{*}{\textbf{Method}} & \multicolumn{2}{c|}{\textbf{Batch Size=1}} & \multicolumn{2}{c|}{\textbf{Batch Size=2}} & \multicolumn{2}{c|}{\textbf{Batch Size=4}} & \multicolumn{2}{c|}{\textbf{Batch Size=8}} & \multicolumn{2}{c|}{\textbf{Batch Size=16}} & \multicolumn{2}{c}{\textbf{Batch Size=32}} \\
    & & \textbf{TTFT(s)} & \textbf{TPOT(s)} & \textbf{TTFT(s)} & \textbf{TPOT(s)} & \textbf{TTFT(s)} & \textbf{TPOT(s)} & \textbf{TTFT(s)} & \textbf{TPOT(s)} & \textbf{TTFT(s)} & \textbf{TPOT(s)} & \textbf{TTFT(s)} & \textbf{TPOT(s)} \\
    \midrule
    \multirow{3}{*}{MMMU} & baseline & 2.810 & 0.786 & 3.819 & 0.853 & 5.100 & 0.981 & 7.949 & 1.414 & 11.562 & 2.078 & 17.626 & 2.552 \\
    & SMoES & \textbf{2.519} & \textbf{0.703} & \textbf{3.303} & \textbf{0.767} & \textbf{4.302} & \textbf{0.890} & \textbf{6.203} & \textbf{1.287} & \textbf{8.804} & \textbf{1.899} & \textbf{13.638} & \textbf{2.309} \\
    & $\Delta$ & \textcolor{darkgreen}{↓10.3\%} & \textcolor{darkgreen}{↓10.5\%} & \textcolor{darkgreen}{↓13.5\%} & \textcolor{darkgreen}{↓10.1\%} & \textcolor{darkgreen}{↓15.7\%} & \textcolor{darkgreen}{↓9.3\%} & \textcolor{darkgreen}{↓22.0\%} & \textcolor{darkgreen}{↓9.0\%} & \textcolor{darkgreen}{↓23.9\%} & \textcolor{darkgreen}{↓8.6\%} & \textcolor{darkgreen}{↓22.6\%} & \textcolor{darkgreen}{↓9.6\%} \\
    \midrule
    \multirow{3}{*}{SQA-IMG} & baseline & 1.493 & 0.766 & 2.940 & 0.858 & 4.217 & 0.951 & 5.824 & 1.278 & 9.759 & 2.100 & 15.036 & 2.721 \\
    & SMoES & \textbf{1.356} & \textbf{0.692} & \textbf{2.558} & \textbf{0.775} & \textbf{3.656} & \textbf{0.858} & \textbf{4.859} & \textbf{1.134} & \textbf{7.571} & \textbf{1.885} & \textbf{12.023} & \textbf{2.431} \\
    & $\Delta$ & \textcolor{darkgreen}{↓9.2\%} & \textcolor{darkgreen}{↓9.7\%} & \textcolor{darkgreen}{↓13.0\%} & \textcolor{darkgreen}{↓9.7\%} & \textcolor{darkgreen}{↓13.3\%} & \textcolor{darkgreen}{↓9.8\%} & \textcolor{darkgreen}{↓16.6\%} & \textcolor{darkgreen}{↓11.3\%} & \textcolor{darkgreen}{↓22.4\%} & \textcolor{darkgreen}{↓10.3\%} & \textcolor{darkgreen}{↓20.0\%} & \textcolor{darkgreen}{↓10.6\%} \\
    \bottomrule
    \end{tabular}
\end{table*}

\begin{figure*}[t]
    \centering
    \includegraphics[width=0.8\textwidth]{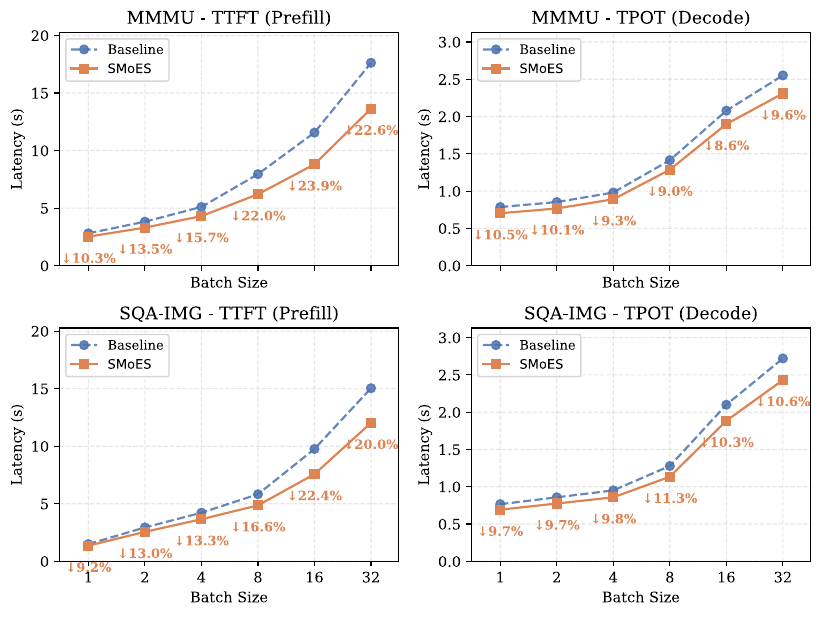}
    \caption{
        Latency decrease of SMoES compared to soft routing baseline at different batch sizes on edge-side deployment.
        The deployed VLM is based on OLMoE with 7B parameters.
        TTFT: Time to First Token (prefill stage); TPOT: Time Per Output Token (decode stage).
    }
    \label{fig:suppl-deployment-speed}
\end{figure*}

We deploy our experiments on a dual-NVIDIA-Orin GPU setup, which represents a typical edge-side scenario for automotive applications. 
Given the high efficiency requirements for edge-side deployment, we test with a small-scale OLMoE-based VLM model with 7B parameters. 
Two NVIDIA Orin GPUs are connected via 10Gb Ethernet and deployed with Expert Parallelism (EP). 
Memory capacity is one of the bottlenecks for edge-side resources. 
Compared to Tensor Parallel (TP) and Data Parallel (DP) strategies, this approach has no redundancy in either weights or KV Cache. 
Compared to Attention-FFN separation with the same memory resource redundancy, this deployment has lower token transfer time, and memory space usage is relatively balanced.
For the baseline deployment, since the experts are balanced, asynchronous transmission cannot give performance benefits and may even reduce efficiency, so synchronous transmission is adopted.
For our \method{} with modality specialization, since there are more local experts with longer computation time, we adopted asynchronous transmission, allowing network transmission and expert computation to execute in parallel.
Additionally, the optimized deployment is similar to PD separation, where KV Cache transmission can be transferred asynchronously and be overlapped.
The baseline uses a sequential order to assign experts, which is equivalent to a random order since there are no order constraints during training.

Based on the previous discussion, we can see that experts exhibit significant specialization. 
During actual deployment and data transmission, the data goes through DeDup (de-duplication) to avoid transmitting duplicate tokens routed to experts on the same device in top-K routing. 
As shown in \cref{tab:suppl-deployment-communication}, we report the cross-GPU EP transfer ratio for vision and text tokens separately, since the number of vision and text tokens differs between prefill and decode phases. 
We also provide the overall proportion of transmitted tokens (V+T) at prefill stage. 
The table compares the transfer ratios with and without DeDup, demonstrating that our specialized methods (SMoES) significantly reduce the communication overhead for most of the cases, especially for vision tokens during the prefill stage.
The conclusion is consistent among different benchmark datasets, indicating that our specialized methods (SMoES) are effective in reducing the communication overhead for most of the vision-language tasks.

The inference TTFT (Time to First Token) (Prefill) time and TPOT (Time Per Output Token) (Decode) time are reported in \cref{tab:suppl-deployment-speed} and visualized in \cref{fig:suppl-deployment-speed}. 
The performance improvement mainly comes from the reduction in communication time and the parallel execution of expert computation and communication, as evidenced by the latency breakdown in \cref{fig:suppl-deployment-speed}. 
For Prefill, as the batch size increases, the proportion of communication time increases, resulting in significant optimization performance, which aligns with the widening gaps observed in both \cref{tab:suppl-deployment-speed,fig:suppl-deployment-speed}. 
For Decode, when the Batch Size is small, due to the limited number of tokens, fewer experts are activated. 
Since the activated experts are proportional to the network transmission volume to a certain extent, the proportion of network transmission time remains constant, and thus the improvement ratio also remains constant.

\subsubsection{Larger-Scale EP}

Edge-side VLM/VLA deployment is crucial for autonomous driving and robotics. 
Scaling SMoES to cloud environments presents complex optimization challenges—where scheduling must account for multidimensional variables such as query modality ratios, expert specialization/redundancy, and scale-up/out bandwidth—yet it also unlocks greater efficiency potential. 
With proper optimization, SMoES's specialized expert groups allow converting costly cross-node communication into local computation, significantly reducing overhead compared to modality-agnostic baselines. 
While fully solving this complex scheduling is non-trivial, we evaluate a 16-GPU (2-node) scenario using a simple modality-based arrangement to reveal the substantial potential of SMoES. 
Without deep optimization, SMoES still reduces the inference time by 5.6\% compared to the baseline (\cref{tab:suppl-larger-scale-ep}).

\begin{table}[ht]
    \centering
    \footnotesize
    \begin{tabular}{lcccccc}
        \toprule
        Batch size & 1 & 2 & 4 & 8 & 16 & 32 \\
        \midrule
        Baseline & 4.3 & 7.2 & 13.2 & 24.1 & 47.3 & 93.4 \\
        SMoES & 4.1 & 6.7 & 12.1 & 23.1 & 45.4 & 88.2 \\
        $\Delta$ & -4.7\% & -6.9\% & -8.3\% & -4.2\% & -4.0\% & -5.6\% \\
        \bottomrule
    \end{tabular}
    \caption{Inference time (s) of Qwen3-MoE (16 GPUs on 2 nodes). }
    \label{tab:suppl-larger-scale-ep}
\end{table}

\subsubsection{Training Overhead}
The training overhead is shown in \cref{tab:suppl-training-overhead}.
SMoES does not affect model parameters, except that the Gaussian Estimator adds a small buffer of approximately 60M. 
TFLOPS and peak GPU memory usage are almost unchanged. 
SMoES introduces a slight increase in training time, with the Gaussian Estimator contributing more due to global feature synchronization. 
But it becomes less noticeable as the model scales up. 
Note that SMoES introduces no additional overhead during inference.

\begin{table}[ht]
    \centering
    \footnotesize
    \setlength{\tabcolsep}{2pt}
    \begin{tabular}{lcccc}
        \toprule
        Model & Params (B) & TFLOPS & Step Time (s) & Mem. (GB)  \\
        \midrule
        32 experts & 1.19 & 1.82 & 0.74 & 22.3 \\
        +inter-bin MI-loss & 1.19 & 1.82 & 0.80 & 22.3 \\
        +Attention Estimator & 1.19 & 1.82 & 0.81 & 22.3 \\
        +Gaussian Estimator & 1.25 & 1.83 & 0.93 & 22.3 \\
        \midrule
        64 experts & 1.59 & 1.90 & 0.82 & 31.4 \\
        +inter-bin MI-loss & 1.59 & 1.90 & 0.90 & 31.4 \\
        +Attention Estimator & 1.59 & 1.90 & 0.91 & 31.4 \\
        +Gaussian Estimator & 1.65 & 1.91 & 0.99 & 31.4 \\
        \bottomrule
    \end{tabular}
    \caption{Training costs on OLMoE for one GPU (batch 2, EP.8). }
    \label{tab:suppl-training-overhead}
\end{table}

\subsection{Visualizations}

\begin{figure*}[t]
\centering
\includegraphics[width=0.99\textwidth]{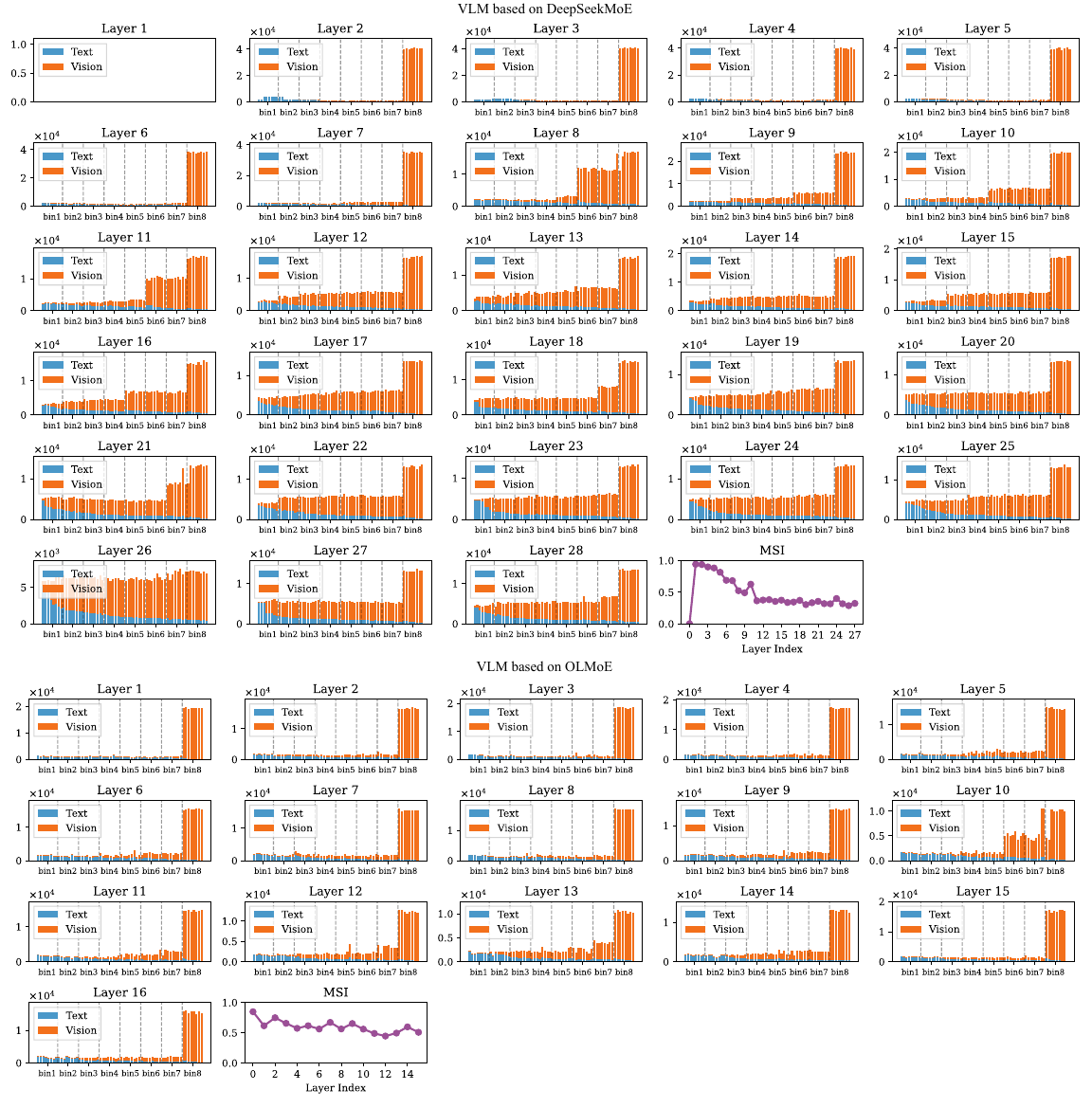}
\caption{
    Routing distribution of tokens to experts in DeepSeekMoE and OLMoE.
    Horizontal axis: all experts grouped into eight expert bins.
    Vertical axis: number of tokens routed to each expert.
}
\label{fig:sup-routing-ds-ol}
\end{figure*}

\begin{figure*}[t]
\centering
\includegraphics[width=0.99\textwidth]{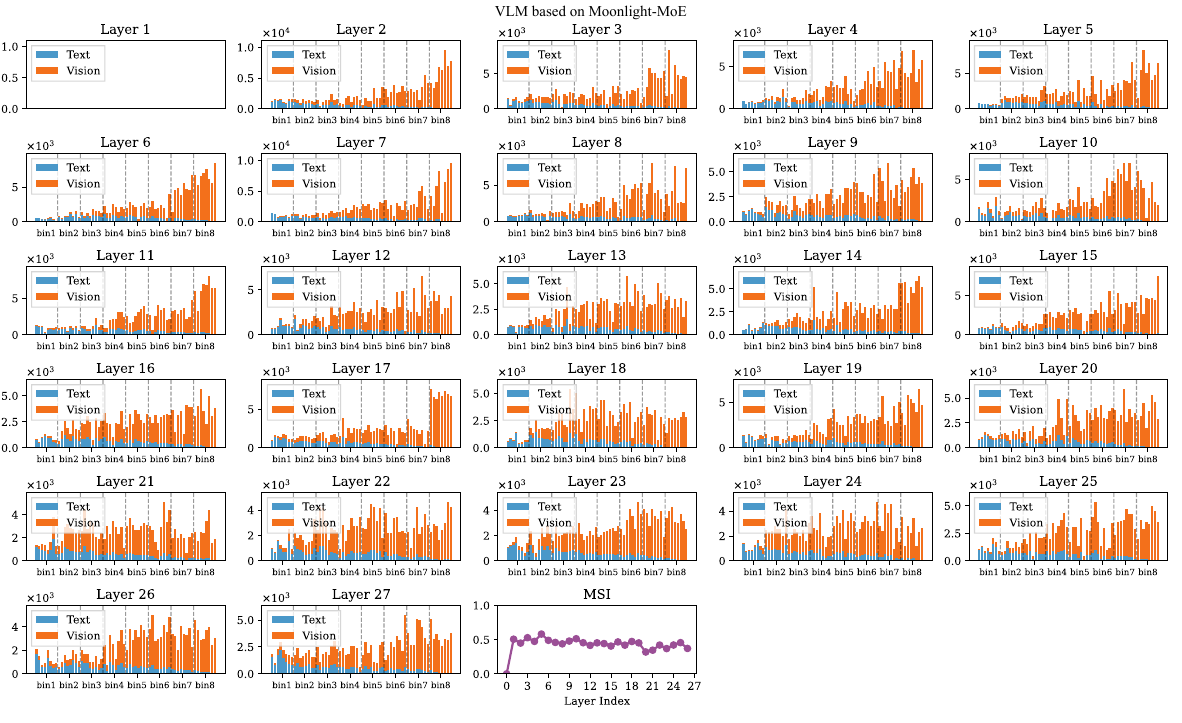}
\caption{
    Routing distribution of tokens to experts in Moonlight-MoE.
    Horizontal axis: all experts grouped into eight expert bins.
    Vertical axis: number of tokens routed to each expert.
}
\label{fig:sup-routing-ml}
\end{figure*}

\begin{figure*}[t]
\centering
\includegraphics[width=0.99\textwidth]{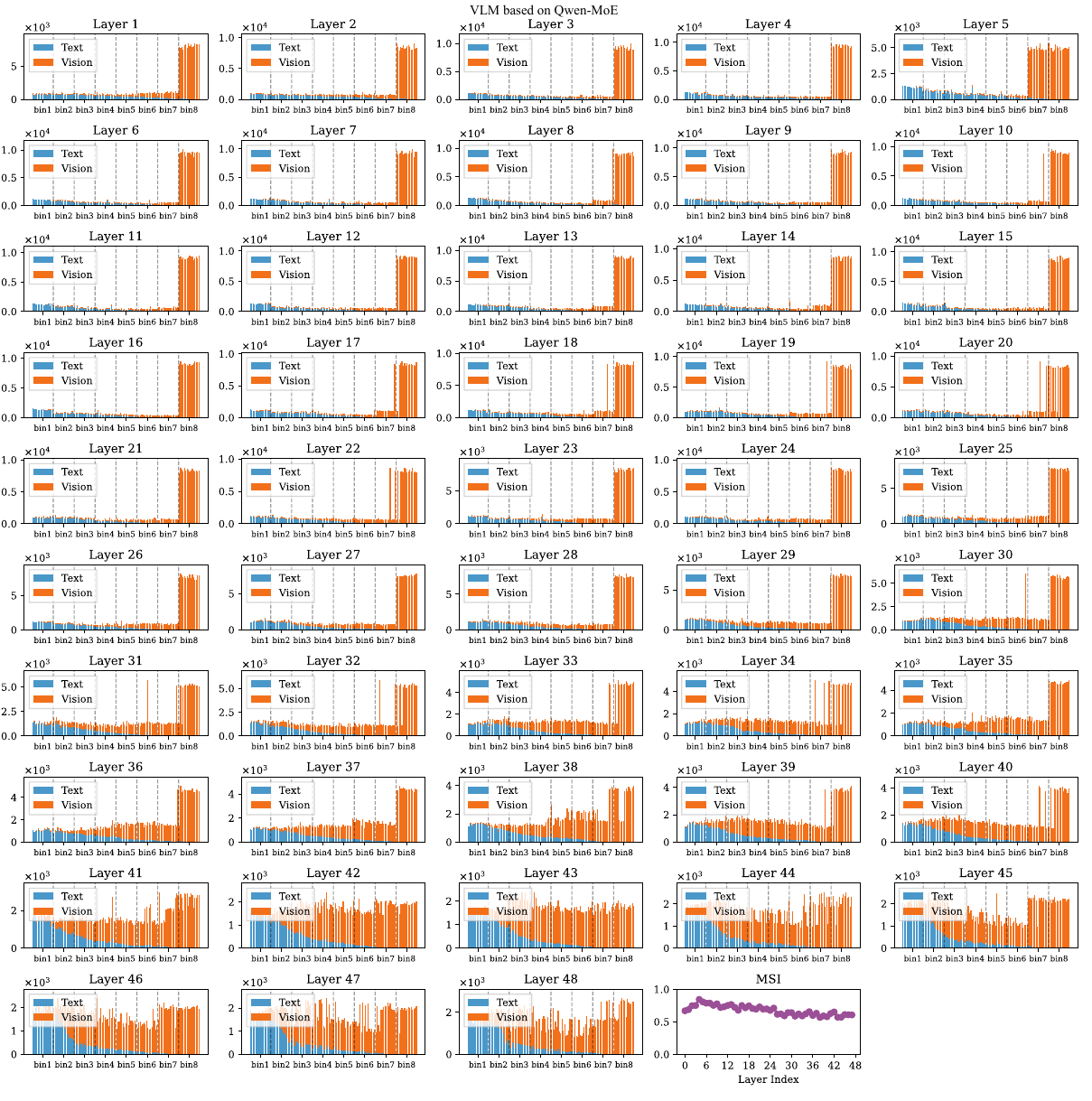}
\caption{
    Routing distribution of tokens to experts in Qwen3-MoE.
    Horizontal axis: all experts grouped into eight expert bins.
    Vertical axis: number of tokens routed to each expert.
}
\label{fig:sup-routing-qw}
\end{figure*}

We show routing distributions for \method{} based on the four backbones in \cref{fig:sup-routing-ds-ol,fig:sup-routing-ml,fig:sup-routing-qw}.
We can see that at most layers, there is one or several bins that are dedicated to vision tokens. 
In vision-language tasks, vision tokens typically dominate in quantity but have lower information density. 
By allocating dedicated experts to handle these abundant yet information-sparse vision tokens, other experts are freed up to process more complex text information. 
This is why SMoES achieves significantly better performance on text-only tasks compared to other methods.

However, vision tokens are not always low in information density; some vision tokens carry critical information. 
Fortunately, our modality specialization method is soft and dynamic, allowing the model to learn to route different vision tokens to different experts. 
This enables important vision tokens to be assigned to specialized experts for processing, preventing information loss. 
This is why our method outperforms manually pre-defined hard-routed and hybrid-routed approaches.

Except for Moonlight-MoE, which shows a relatively flat MSI layer-wise curve, most models exhibit the characteristic that shallow layers have higher modality specialization (MSI) while deeper layers show lower specialization, indicating a gradual fusion of modalities as the network goes deeper. 
For Moonlight-MoE, the additional router score bias makes its routing distribution less uniform, which slightly reduces its overall modality specialization level compared to other backbones.

\begin{figure*}[t]
\centering
\includegraphics[width=0.99\textwidth]{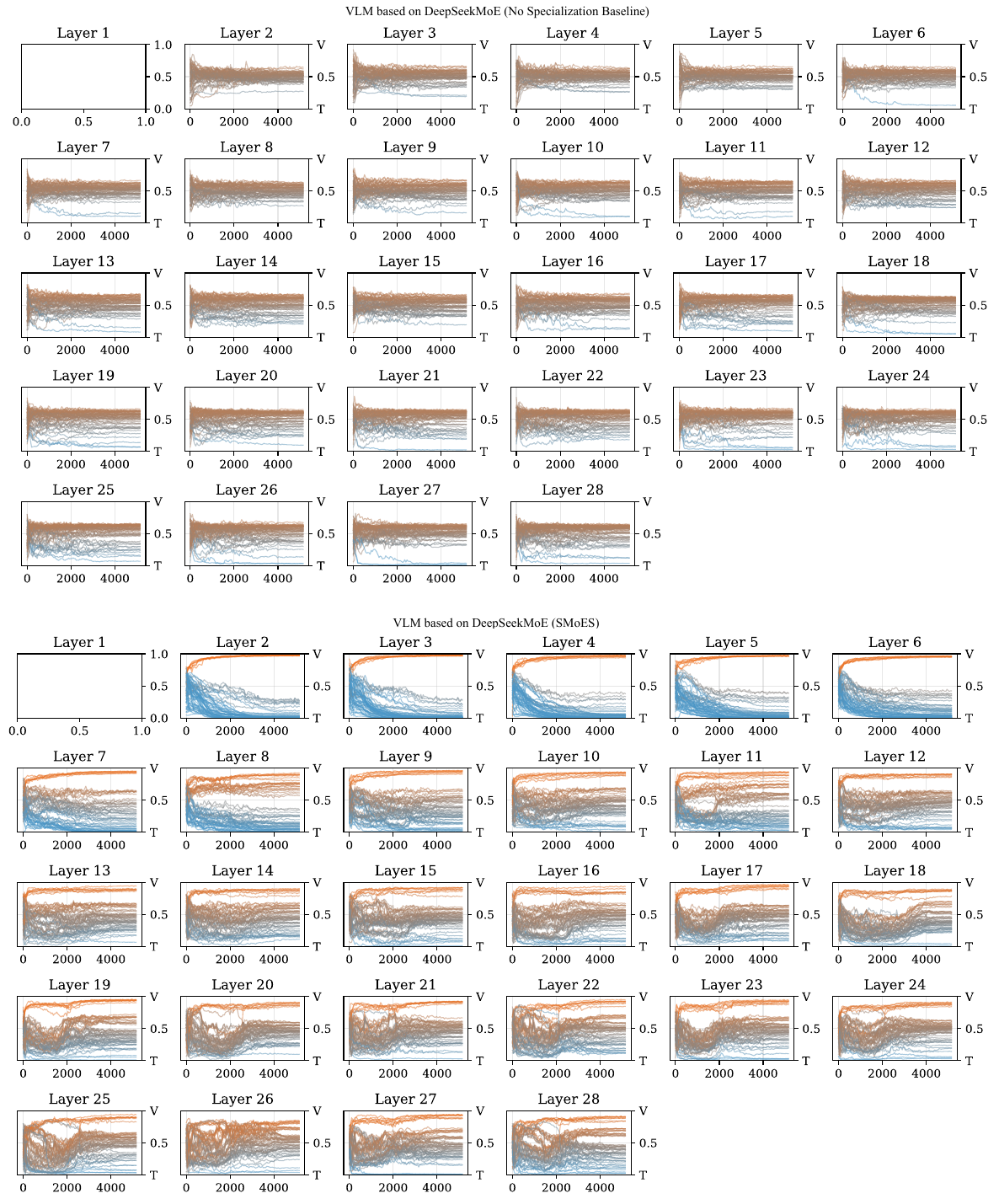}
\caption{
    Evolution of expert specialization during training on DeepSeekMoE.
    Each curve represents an expert.
    Horizontal axis: training steps.
    Vertical axis: expert specialization score (symmetric expansion of MSI).
    V: vision specialization; T: text specialization.
}
\label{fig:sup-specialization-ds}
\end{figure*}

\begin{figure*}[t]
\centering
\includegraphics[width=0.99\textwidth]{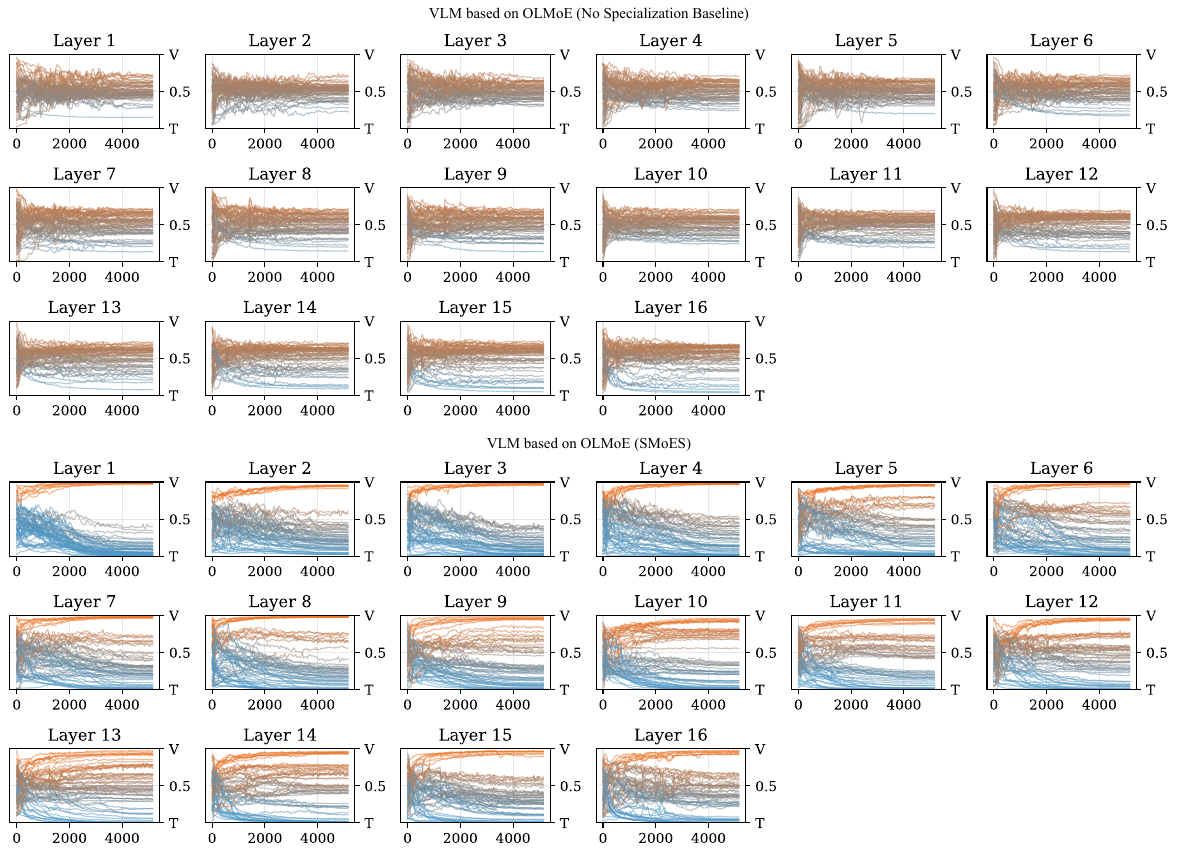}
\caption{
    Evolution of expert specialization during training on OLMoE.
    Each curve represents an expert.
    Horizontal axis: training steps.
    Vertical axis: expert specialization score (symmetric expansion of MSI).
    V: vision specialization; T: text specialization.
}
\label{fig:sup-specialization-ol}
\end{figure*}

\begin{figure*}[t]
\centering
\includegraphics[width=0.99\textwidth]{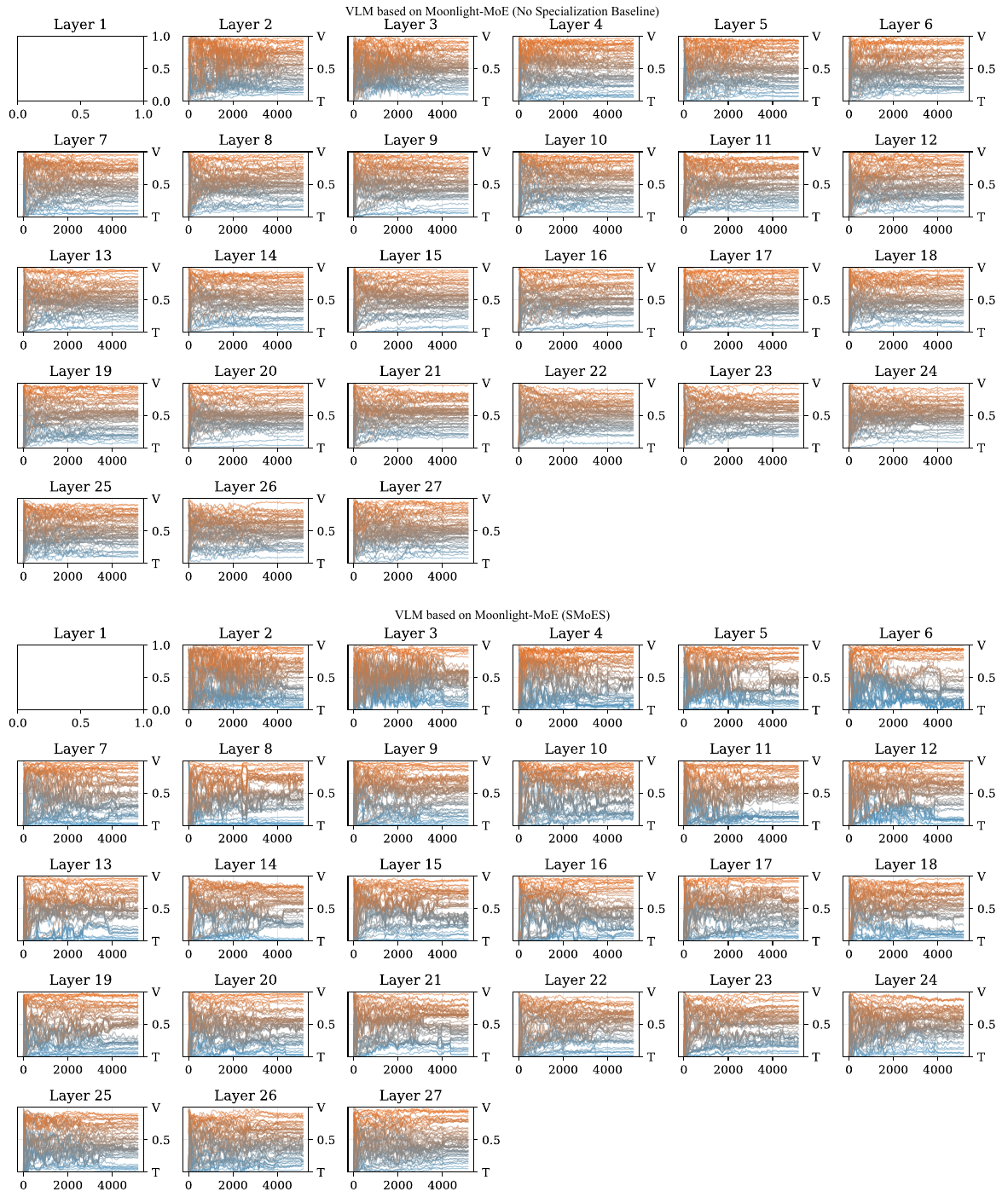}
\caption{
    Evolution of expert specialization during training on Moonlight-MoE.
    Each curve represents an expert.
    Horizontal axis: training steps.
    Vertical axis: expert specialization score (symmetric expansion of MSI).
    V: vision specialization; T: text specialization.
}
\label{fig:sup-specialization-ml}
\end{figure*}

\begin{figure*}[t]
\centering
\includegraphics[width=0.99\textwidth]{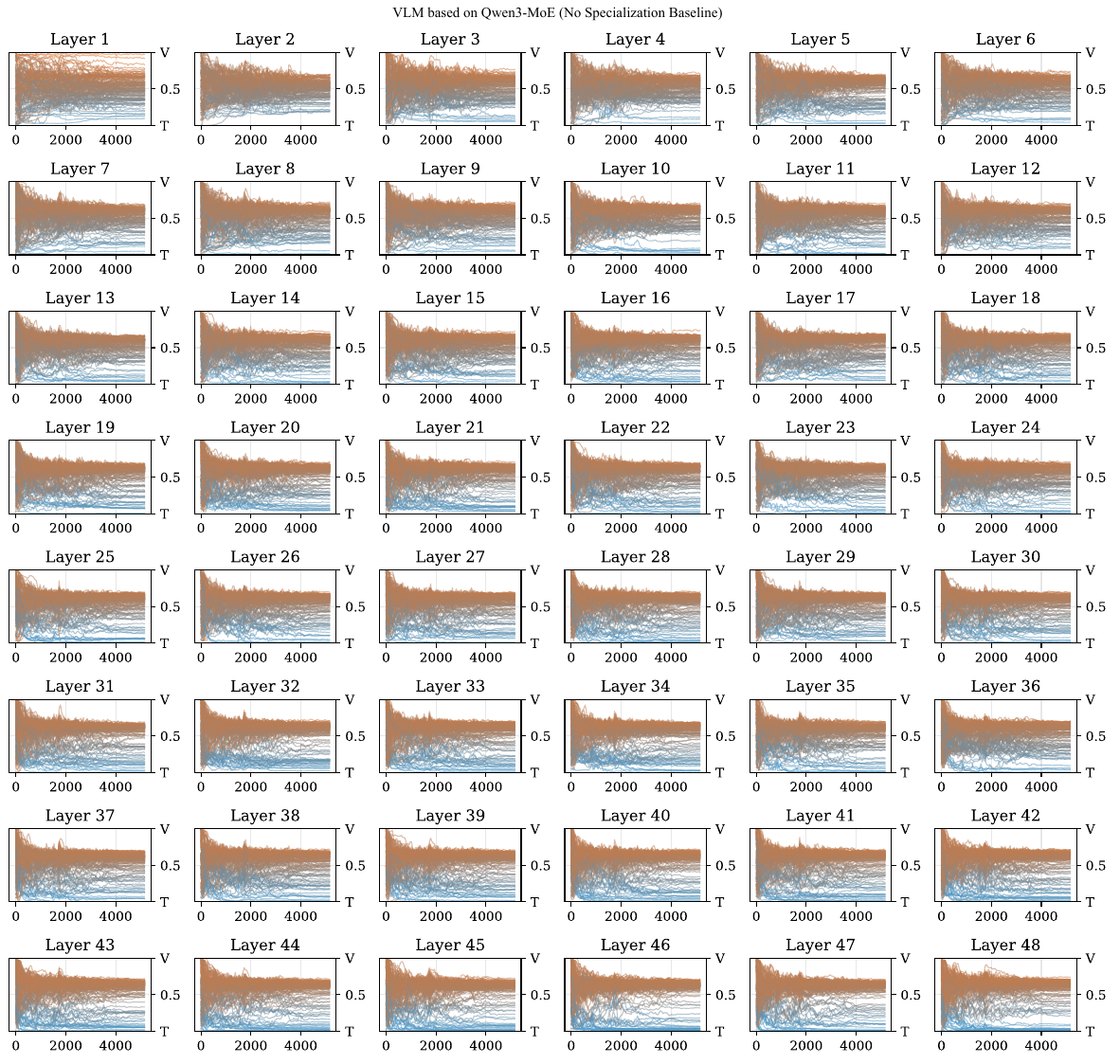}
\caption{
    Evolution of expert specialization during training on Qwen3-MoE with no specialization (baseline).
    Each curve represents an expert.
    Horizontal axis: training steps.
    Vertical axis: expert specialization score (symmetric expansion of MSI).
    V: vision specialization; T: text specialization.
}
\label{fig:sup-specialization-qw-baseline}
\end{figure*}

\begin{figure*}[t]
\centering
\includegraphics[width=0.99\textwidth]{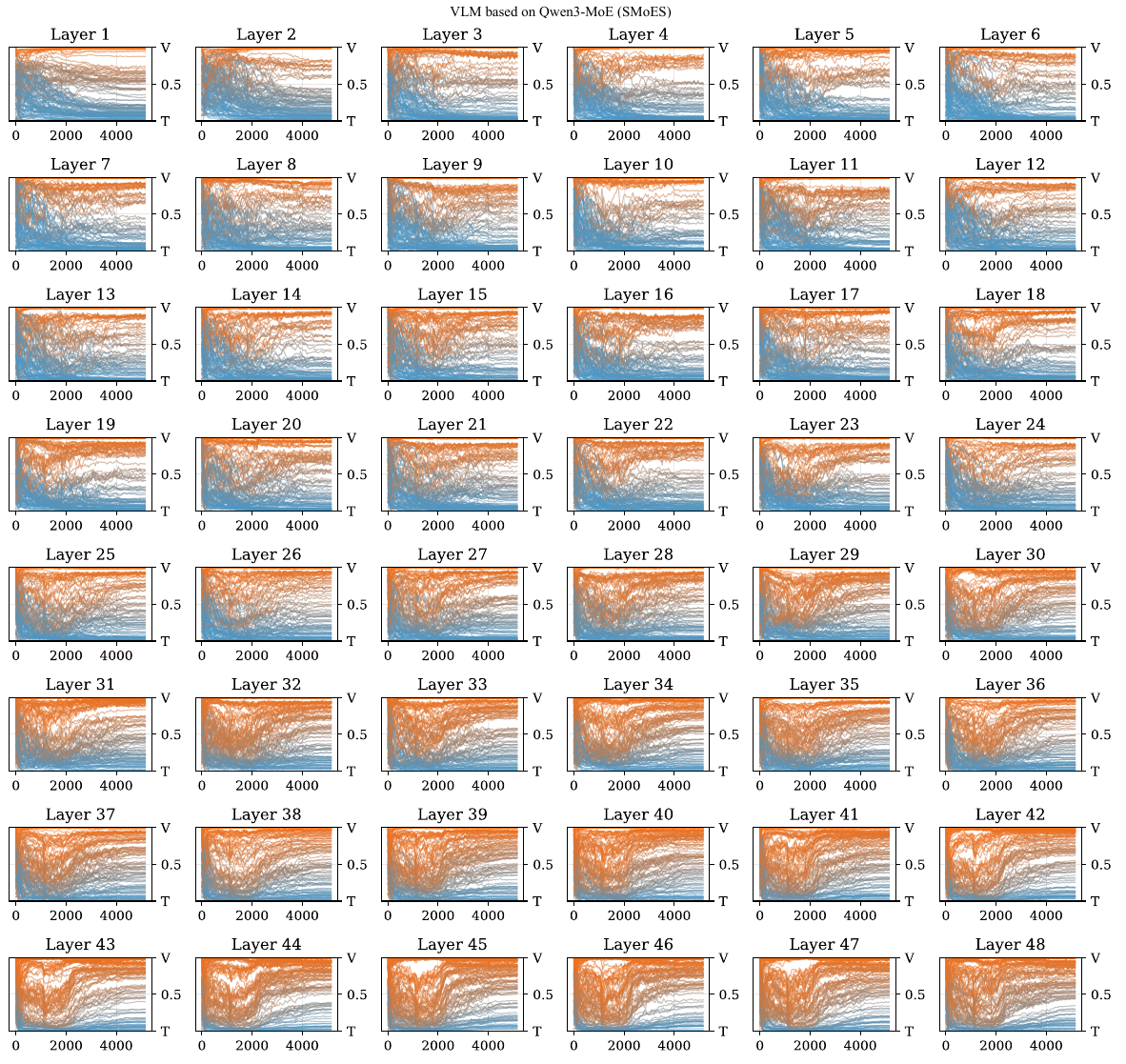}
\caption{
    Evolution of expert specialization during training on Qwen3-MoE with SMoES.
    Each curve represents an expert.
    Horizontal axis: training steps.
    Vertical axis: expert specialization score (symmetric expansion of MSI).
    V: vision specialization; T: text specialization.
}
\label{fig:sup-specialization-qw-smoes}
\end{figure*}

We show the evolution of expert specialization during training on the four backbones in \cref{fig:sup-specialization-ds,fig:sup-specialization-ol,fig:sup-specialization-ml,fig:sup-specialization-qw-baseline,fig:sup-specialization-qw-smoes}.
At the early stage of training, the baseline and SMoES exhibit similar expert tendencies, with specialization scores roughly evenly spread across the spectrum.
However, as training progresses, baseline experts tend to collapse toward the middle, meaning most experts become mixed-modal and handle both vision and text tokens simultaneously.
In contrast, SMoES maintains pronounced modality specialization: one subset of experts separates to process vision-dominant tokens, another focuses on text-dominant tokens, and the remaining experts stay mixed to handle multimodal inputs.
This specialized pattern is especially evident in shallow layers where vision and text tokens remain far apart in representation space; deeper layers, after repeated cross-modal interactions, naturally become more multimodal, which explains the reduced but still observable specialization gap.

\begin{figure*}[t]
    \centering
    \includegraphics[width=0.99\textwidth]{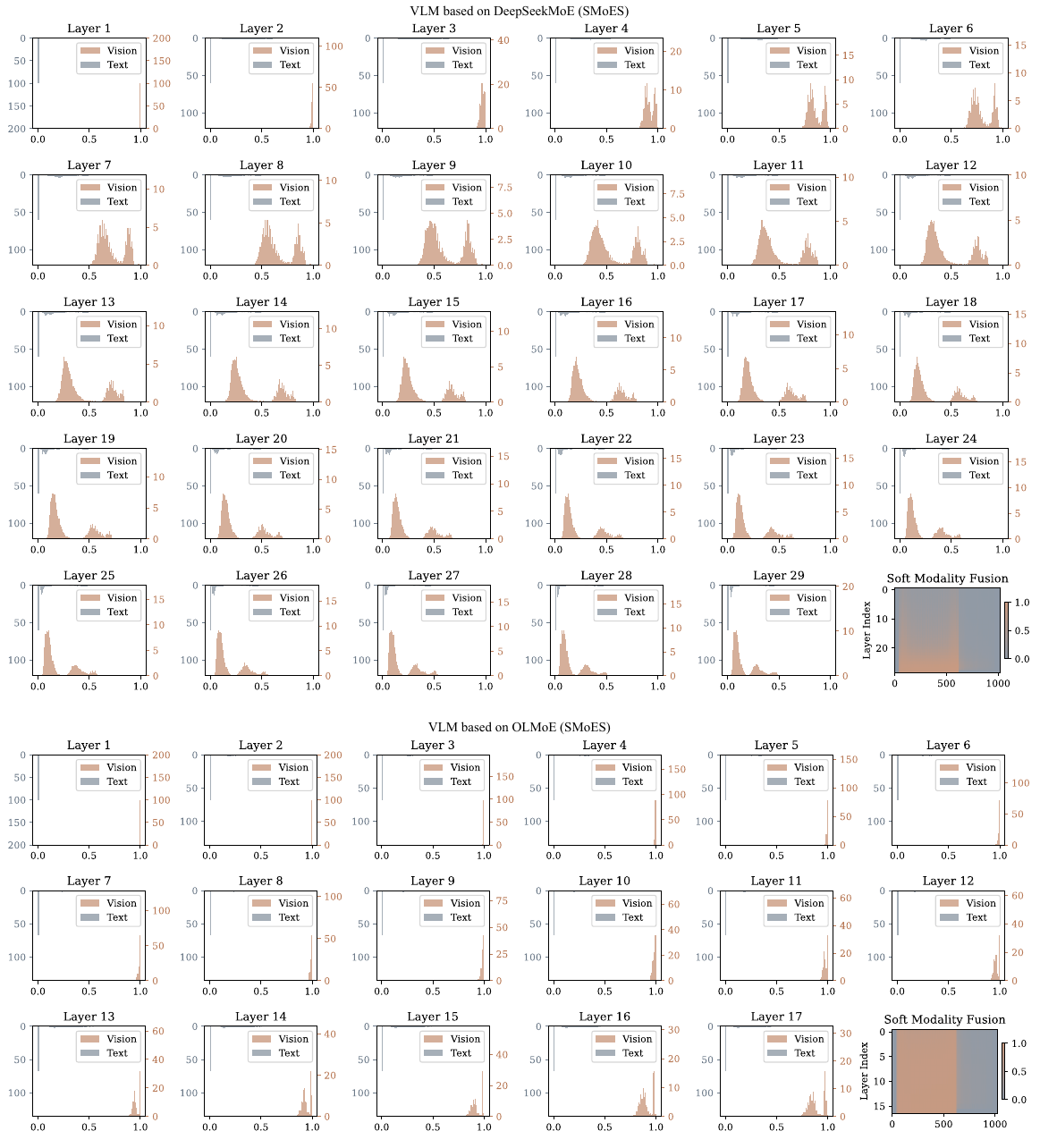}
    \caption{
        Modality fusion patterns for $\method{}_\text{attention-soft}$ in DeepSeekMoE and OLMoE.
        Layer subplots: horizontal axis is the soft modality score, and vertical axis is the number of token distribution.
        Soft modality fusion subplot: horizontal axis is the token id in a sample sequence, and vertical axis is the layer index.
    }
    \label{fig:sup-modality-fusion-ds-ol}
\end{figure*}

\begin{figure*}[t]
    \centering
    \includegraphics[width=0.99\textwidth]{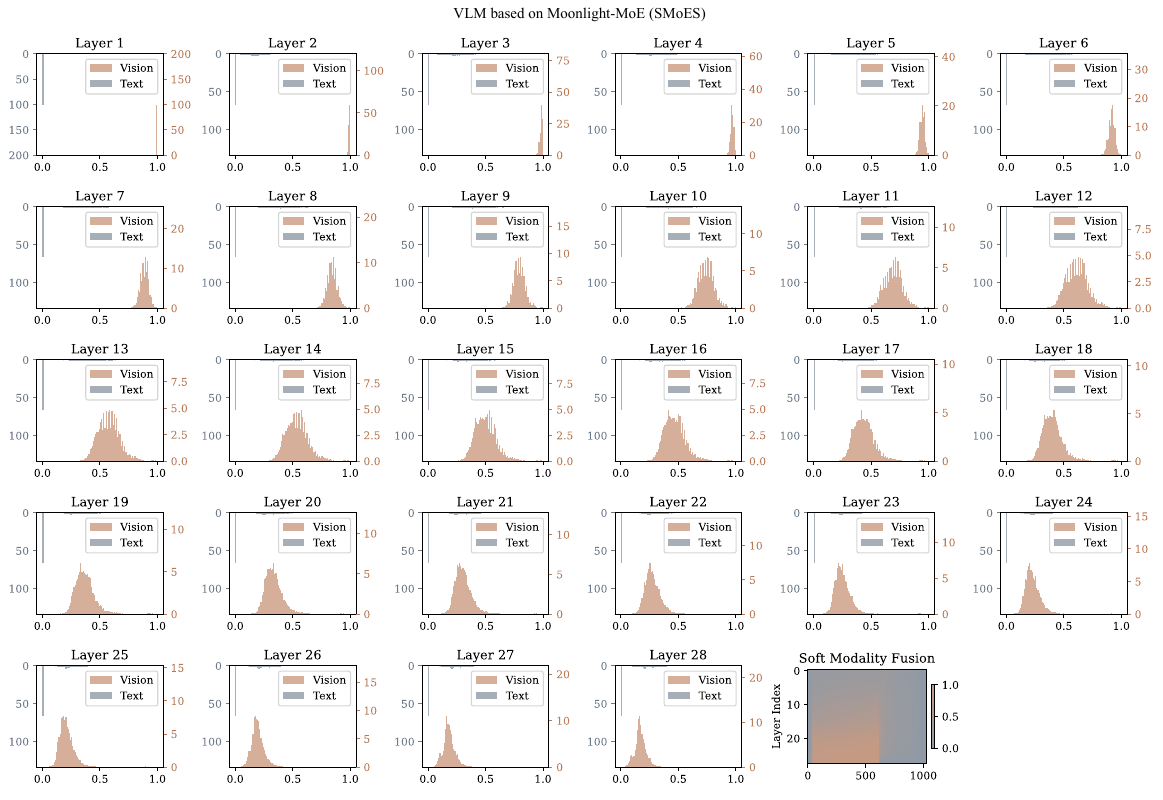}
    \caption{
        Modality fusion patterns for $\method{}_\text{attention-soft}$ in Moonlight-MoE.
        Layer subplots: horizontal axis is the soft modality score, and vertical axis is the number of token distribution.
        Soft modality fusion subplot: horizontal axis is the token id in a sample sequence, and vertical axis is the layer index.
    }
    \label{fig:sup-modality-fusion-ml}
\end{figure*}

\begin{figure*}[t]
    \centering
    \includegraphics[width=0.99\textwidth]{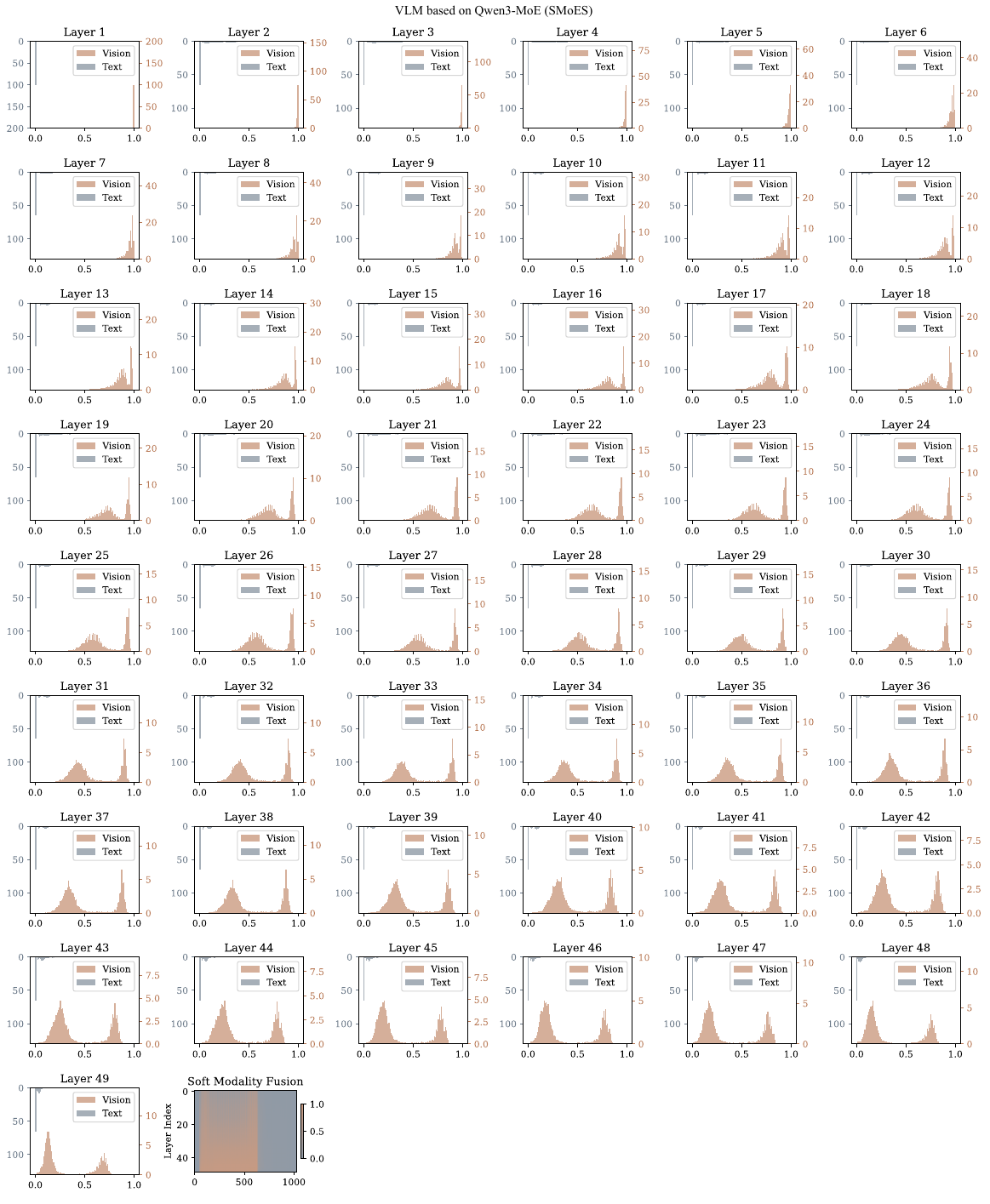}
    \vspace{-0.5em}
    \caption{
        Modality fusion patterns for $\method{}_\text{attention-soft}$ in Qwen3-MoE.
        Layer subplots: horizontal axis is the soft modality score, and vertical axis is the number of token distribution.
        Soft modality fusion subplot: horizontal axis is the token id in a sample sequence, and vertical axis is the layer index.
    }
    \label{fig:sup-modality-fusion-qw}
\end{figure*}

We show the modality fusion patterns for $\method{}_\text{attention-soft}$ in the four backbones in \cref{fig:sup-modality-fusion-ds-ol,fig:sup-modality-fusion-ml,fig:sup-modality-fusion-qw}.
Across models, shallow layers tend to produce soft modality scores near the two extremes (pure vision or pure text), and the scores gradually converge toward the center as depth increases.
The convergence rate and shape vary with the backbone: DeepSeekMoE and Qwen3-MoE often split vision tokens into multiple peaks, Moonlight-MoE keeps a single peak.
These differences indicate that backbone architecture and pretraining induce distinct token characteristics, while also showing that SMoES adapts to diverse MoE designs.

\end{document}